\documentclass{article}
\usepackage[main,final]{neurips_2026}
\usepackage[utf8]{inputenc}
\usepackage[T1]{fontenc}
\usepackage{amssymb}
\usepackage{amsmath}
\usepackage{amsbsy}
\usepackage{amsfonts}
\usepackage[nice]{nicefrac}
\usepackage{breqn}
\usepackage{mathtools}
\usepackage{graphicx}
\usepackage{subcaption}
\usepackage{svg}
\usepackage{tikz}
\usepackage{tikz-cd}
\usepackage{pgfplots}
\usepackage{float}
\usepackage{dblfloatfix}
\usepackage{hyperref}
\hypersetup{hidelinks}
\usepackage{booktabs}
\usepackage{microtype}
\usepackage{multirow}
\usepackage{booktabs}
\usepackage{adjustbox}
\usepackage{soul}
\usepackage{tcolorbox}
\tcbuselibrary{skins}
\usepackage[dvipsnames]{xcolor}
\allowdisplaybreaks
\pgfplotsset{compat=1.16}
\newcommand\blfootnote[1]{%
  \begingroup
  \renewcommand\thefootnote{}\footnotetext{#1}%
  \addtocounter{footnote}{0}%
  \endgroup
}
\usepackage{array}
\usepackage{calc}
\newcolumntype{C}[1]{>{\centering\arraybackslash}m{#1}}
\newlength{\compcolw}
\definecolor{PrincetonOrange}{HTML}{E77500}
\title{Discovering Phase Space Structure in Learned Hamiltonian Systems}
\author{%
    Yulong Yang$^{*,\dagger}$ \quad Jiayin Liu$^{*}$ \quad Vineet Bansal \quad Christine Allen-Blanchette$^{\dagger}$\\
    Princeton University\\
    \texttt{\{yulong.yang, jl4266, vineetb, ca15\}@princeton.edu}\\
}
\begin{document}
\maketitle
\blfootnote{$^{*}$ Equal contribution. $^{\dagger}$ Corresponding author. This work was done while Jiayin Liu was a visiting student at Princeton.} 
\begin{abstract}
Mechanics-inspired neural networks provide physically structured models of dynamical systems, but typically assume access to the coordinates in which the underlying dynamics are expressed. While recent generative approaches relax this assumption, they remain limited in their ability to jointly model different dynamical regimes and discover interpretable latent structure. We introduce Structured Phase Space GAN (SPS-GAN), a conditional generative framework that learns mechanical dynamics and discovers latent phase-space structure from Cartesian trajectories or video observations. We achieve this by combining adversarial generation with port-Hamiltonian latent dynamics and extending cyclic-coordinate regularization to account for dissipation and forcing. This formulation supports conservative, dissipative, and forced systems while encouraging compact latent representations from which active degrees of freedom can be identified. Experiments on simulated and real-world systems demonstrate improved trajectory accuracy and video generation quality over competitive baselines, while identifying active latent dimensions consistent with the expected effective degrees of freedom and recovering physically interpretable coordinates across observation modalities and dynamical regimes.
\end{abstract}
\section{Introduction}
Learning models of dynamical systems from observations is fundamental to applications in robotics and control~\citep{khalil2002nonlinear}, autonomous driving~\citep{lefevre2014survey}, and aerospace engineering~\citep{o2024reconstructed}.
However, learning physically interpretable models remains challenging because the generalized coordinates that describe the underlying mechanical structure are often not directly observable.
Although observations may consist of Cartesian trajectories or video frames, the underlying motion may admit a lower-dimensional mechanical representation that reflects the system's constraints and symmetries.
Classical system identification methods learn dynamical models from observations~\citep{brunton2016discovering,galioto2020bayesian2,paredes2021identification,wu2024identifying,paredes2024output,richards2024output}, but often rely on prescribed model structures or assumptions about the underlying state representation.
This motivates learning models that not only reproduce observed dynamics but also discover compact, physically interpretable representations of the underlying motion.

Recent neural network approaches have incorporated principles from classical mechanics to improve the physical consistency and interpretability of learned dynamics.
Hamiltonian and Lagrangian neural networks learn energy functions whose induced equations of motion provide structured predictions of system evolution~\citep{greydanus2019hamiltonian,chen2019symplectic,cranmer2020lagrangian}.
Extensions to dissipative and forced systems broaden the class of dynamics that can be modeled~\citep{zhong2020dissipative,sosanya2022dissipative,desai2021port}.
Mechanics-inspired generative models further learn distributions over trajectories or videos using structured latent dynamics, including Hamiltonian Generative Networks~\citep{toth2019hamiltonian} and Hamiltonian GAN~\citep{allen2024hamiltonian}.
In particular, Hamiltonian GAN introduces cyclic-coordinate regularization to encourage compact latent phase-space representations, but is formulated for conservative dynamics without conditional generation across systems.
Complementary approaches identify conserved quantities and symmetries directly from data~\citep{kasim2022constants,matsubara2022finde,lishkova2023discrete}, but generally focus on identifying dynamical invariants rather than learning generative models.

Despite these advances, learning generative models that capture physically plausible motion while discovering interpretable latent mechanical structure remains challenging.
We introduce Structured Phase Space GAN (SPS-GAN), a conditional generative framework that addresses this challenge using Cartesian trajectories or video observations.
The central idea is to constrain the generative process through structured latent dynamics, enabling mechanical structure to be identified from the learned phase space.
SPS-GAN maps sampled motion codes and system information to initial latent phase-space states, which evolve according to port-Hamiltonian dynamics. The resulting trajectories are decoded into the observation space.
The port-Hamiltonian formulation supports conservative, dissipative, and forced dynamics, while adversarial training encourages the generated motions to match the observed distribution.
To encourage compact latent structure, we adapt cyclic-coordinate regularization~\citep{allen2024hamiltonian} to account for dissipation and forcing, encouraging redundant coordinate--momentum pairs to become inactive.
Analyzing and decoding the learned latent representation then allows us to identify active dimensions and interpret their relationship to physical mechanical coordinates.

We evaluate SPS-GAN on simulated systems spanning conservative, dissipative, and forced dynamics, as well as real-world trajectory and video datasets.
SPS-GAN achieves competitive or improved trajectory accuracy relative to supervised dynamics baselines and consistently improves video generation quality over generative baselines.
Analysis of the learned phase space further reveals active latent dimensions consistent with the expected effective degrees of freedom and physically interpretable coordinates across the evaluated systems.
Our main contributions are:
\vspace{-2mm}
\begin{itemize}
    \item We introduce SPS-GAN, a conditional generative
    framework for learning physically structured
    dynamics from Cartesian trajectories or video
    observations, supporting generation across
    systems and parameter variations.

    \item We extend Hamiltonian generative modeling
    to dissipative and forced systems through
    port-Hamiltonian latent dynamics, adapting
    cyclic-coordinate regularization to encourage
    compact latent representations.
    \item We demonstrate that SPS-GAN identifies compact latent representations whose active dimensions match the expected effective degrees of freedom of the evaluated systems, with decoded coordinates corresponding to interpretable mechanical motion.
\end{itemize}
\section{Related Work}
%
\textbf{Hamiltonian Mechanics in Neural Networks.}
Hamiltonian inductive biases incorporate principles of classical mechanics into neural dynamical models, providing physically structured representations of system evolution.
Hamiltonian Neural Networks (HNNs)~\citep{greydanus2019hamiltonian} learn a system Hamiltonian and predict dynamics using Hamilton's equations.
Subsequent approaches incorporate symplectic integration, controlled dynamics, and geometric structure to improve physical consistency and long-horizon prediction~\citep{chen2019symplectic,zhong2019symplectic, chen2021data,mattheakis2022hamiltonian,aboussalah2025geohnns}.
Port-Hamiltonian neural networks extend Hamiltonian modeling to dissipative and forced systems by explicitly separating energy-conserving dynamics from dissipation and external forcing~\citep{zhong2020dissipative,desai2021port}.
Mechanics-inspired inductive biases have also been applied to high-dimensional observations, including video prediction of physical dynamics~\citep{allen2020lagnetvip,mason2023learning}.
In contrast to these approaches, SPS-GAN uses port-Hamiltonian latent dynamics within a conditional generative framework to model distributions over physically plausible motions.
%

\textbf{Generating Physically Plausible Data.}
Mechanics-inspired generative models incorporate physical inductive biases to synthesize trajectories and videos with structured dynamics.
Hamiltonian Generative Networks (HGNs)~\citep{toth2019hamiltonian} generate videos using learned latent Hamiltonian dynamics and symplectic integration.
Subsequent approaches explore neural differential equations for video generation~\citep{gordon2021latent} and Hamiltonian latent operators for disentangling motion and appearance~\citep{khan2022hamiltonian}.
Hamiltonian GAN (HGAN)~\citep{allen2024hamiltonian} combines adversarial generation with Hamiltonian latent dynamics to synthesize physically plausible motion.
Beyond generating physically plausible motion, SPS-GAN also discovers compact, interpretable latent mechanical structure from observations.

\textbf{Discovering Structure in Dynamical Systems.}
Complementary approaches seek to discover dynamical structure directly from observations through conserved quantities and symmetries.
COMET~\citep{kasim2022constants} learns candidate constants of motion by enforcing conservation along observed trajectories, while FINDE~\citep{matsubara2022finde} learns first integrals and incorporates them into neural differential equations to preserve the identified invariants during prediction.
Other approaches investigate symmetry structure through prescribed group actions: SymDLNN~\citep{lishkova2023discrete} identifies symmetry subgroups, while Latent MOS~\citep{li2025latent} incorporates predefined symmetry groups into a mixture-of-experts architecture.
These methods focus on identifying invariants or symmetries rather than discovering compact latent representations within a generative model.
Hamiltonian GAN~\citep{allen2024hamiltonian} addresses the latter objective by introducing cyclic-coordinate regularization to encourage compact latent Hamiltonian representations.
SPS-GAN extends this regularization principle to port-Hamiltonian dynamics by accounting for dissipation and forcing, enabling latent structure discovery across conservative and non-conservative systems.
%
\section{Background}
%
We review Hamiltonian mechanics and port-Hamiltonian neural networks, which provide the mathematical foundations for SPS-GAN. In particular, cyclic coordinates motivate the latent structure-discovery regularizer, while port-Hamiltonian dynamics enable modeling of conservative, dissipative, and forced systems.
%
%
\subsection{Hamiltonian Mechanics}\label{subsec:background_hamilton_mechanics}
Hamiltonian mechanics describes conservative mechanical systems in terms of generalized coordinates $\mathbf{q}$ and their conjugate momenta $\mathbf{p}$~\citep{marion2013classical}.
Together, these variables define the phase-space state $\mathbf{z}=(\mathbf{q},\mathbf{p})$.
For the systems considered here, the Hamiltonian $\mathcal{H}(\mathbf{q},\mathbf{p})$ represents the total mechanical energy, and the dynamics follow Hamilton's equations,
\begin{equation}
    \begin{bmatrix}
        \dot{\mathbf{q}}\\ \dot{\mathbf{p}}
    \end{bmatrix} = \begin{bmatrix}
        0 & \mathbf{I}\\ -\mathbf{I} & 0
    \end{bmatrix}\nabla\mathcal{H}(\mathbf{q}, \mathbf{p}).\label{eq:hams_eqns}
\end{equation}
A generalized coordinate $q_k$ is cyclic, or ignorable, if the Hamiltonian is independent of that coordinate, i.e., $\partial\mathcal{H}/\partial q_k=0$.
Hamilton's equations then imply conservation of its conjugate momentum,
\begin{equation}
    \dot{p}_{k} = -\nicefrac{\partial\mathcal{H}}{\partial q_{k}} = 0.
    \label{eqn:cyclic}
\end{equation}
This property motivates the cyclic-coordinate regularization used in Section~\ref{subsec:latent_space_motion_model} to encourage compact latent phase-space representations.
%
%
\subsection{Port Hamiltonian Neural Networks}\label{subsec:background_HNN}
Standard autonomous Hamiltonian dynamics conserve energy and do not directly account for dissipation or external forcing.
The port-Hamiltonian formulation~\citep{maschke1993port} extends Hamiltonian dynamics to incorporate these effects,
\begin{equation}
    \begin{bmatrix}
        \dot{\mathbf{q}}\\ \dot{\mathbf{p}}
    \end{bmatrix} = \left(\begin{bmatrix}
        0 & \mathbf{I}\\ -\mathbf{I} & 0
    \end{bmatrix} - \begin{bmatrix}
        0 & 0\\ 0 & \mathbf{D}
    \end{bmatrix}\right)
        \nabla\mathcal{H}(\mathbf{q}, \mathbf{p}) + \begin{bmatrix} 0\\ \mathbf{G}\left(\mathbf{q}\right)
    \end{bmatrix}\mathbf{u},\label{eqn:port_ham}
\end{equation}
Here, $\mathbf{D}\succeq 0$ captures energy dissipation, $\mathbf{u}$ denotes an external input, and $\mathbf{G}(\mathbf{q})$ maps that input into the momentum dynamics.
This formulation separates energy-conserving dynamics from dissipative effects and external forcing.

For systems with unknown dynamics, Hamiltonian Neural Networks (HNNs)~\citep{greydanus2019hamiltonian} parameterize the Hamiltonian using a neural network and derive the dynamics through Hamilton's equations.
Port-Hamiltonian Neural Networks (pHNNs)~\citep{desai2021port} extend this approach by learning the Hamiltonian alongside dissipation and forcing terms.
%
%
In Section~\ref{subsec:latent_space_motion_model}, we use a pHNN to evolve learned phase-space states, without requiring access to prescribed canonical coordinates.
%
\section{Structured Phase Space GAN}
\begin{figure}[t]
    \centering
    \includegraphics[width=0.9\textwidth]{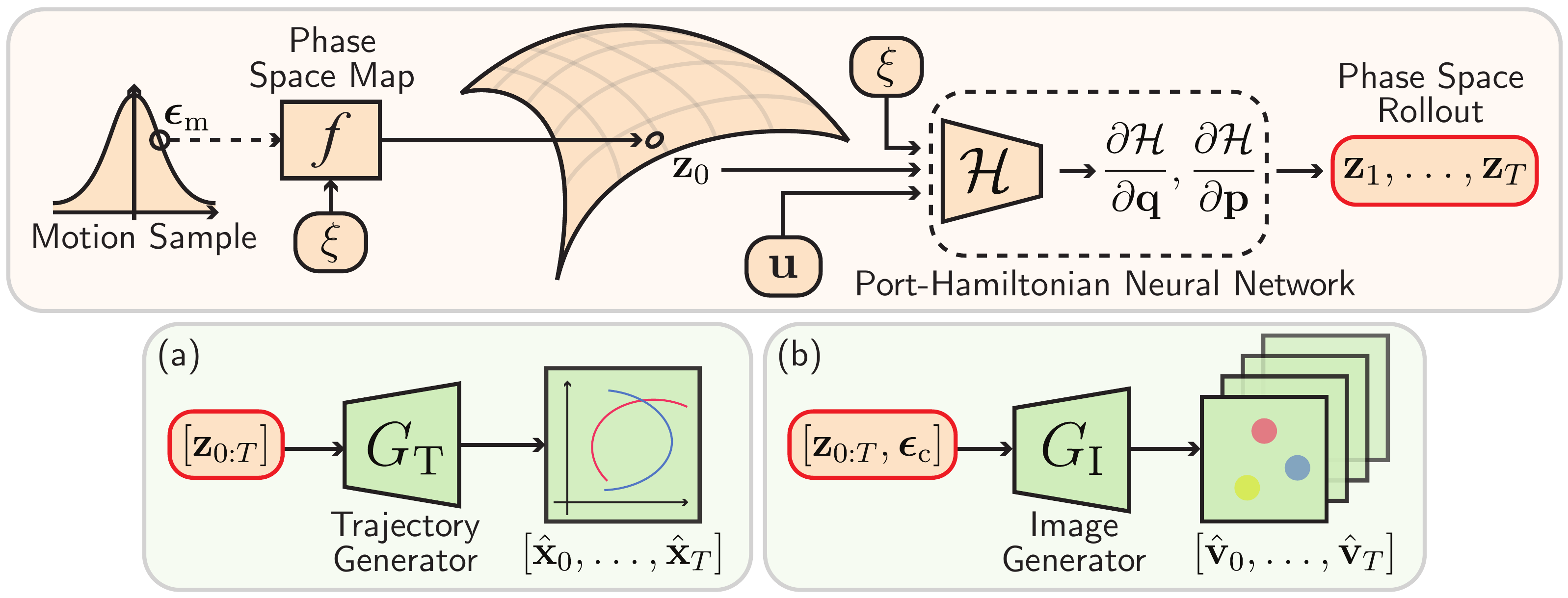}
    \vskip -0.06in
    \caption{\textbf{SPS-GAN generates trajectories and videos across multiple systems through a shared phase-space backbone.} Random motion samples $\boldsymbol{\epsilon}_{\mathrm{m}}$ and system information $\xi$  are passed through the phase-space map $f$ to produce an initial phase-space state $\mathbf{z}_{0}$. This state is evolved forward in time by a Port-Hamiltonian Neural Network~\citep{desai2021port} conditioned on $\xi$, yielding the phase-space trajectory $\left[\mathbf{z}_{0:T}\right]$, which is decoded to \textbf{(a)} Cartesian trajectory $[\hat{\mathbf{x}}_{0:T}]$ and \textbf{(b)} video frames $[\hat{\mathbf{v}}_{0:T}]$.}
    \label{fig:SPSGAN_IntroCartoonCombined}
    \vskip -0.15in
\end{figure}
Structured Phase Space GAN (SPS-GAN) is a conditional generative framework for learning physically structured dynamics from either Cartesian trajectories or video observations, as illustrated in Figure~\ref{fig:SPSGAN_IntroCartoonCombined}.
SPS-GAN maps sampled motion codes and system information to initial latent phase-space states, which evolve according to port-Hamiltonian dynamics before being decoded into observations.
Adversarial training encourages the generated sequences to match the observed data distribution, while cyclic-coordinate regularization encourages compact latent phase-space representations.
\subsection{Phase-Space Map}\label{subsec:config_space}
Port-Hamiltonian dynamics provide a physically structured model of system evolution, but require an initial phase-space state. 
Since SPS-GAN generates trajectories without prescribed initial conditions or canonical coordinates, we learn a conditional map from sampled motion codes to latent initial states.
Let $\xi$ denote system information, such as system identity or physical parameters, and let $\boldsymbol{\epsilon}_{\mathrm{m}}\sim\mathcal{N}\left(0,\mathbf{I}_{d_{\mathrm{m}}}\right)$ denote a random motion code.
An MLP $f$ maps these inputs to an initial latent phase-space state,
\begin{equation}
    \mathbf{z}_{0} = f\left(\boldsymbol{\epsilon}_{\mathrm{m}},\xi\right)
    \in \mathbb{R}^{2 d_{\mathrm{lat}}},
\end{equation}
where $\mathbf{z}_0=(\mathbf{q}_0,\mathbf{p}_0)$ consists of learned generalized coordinates $\mathbf{q}_0\in\mathbb{R}^{d_{\mathrm{lat}}}$ and conjugate momenta $\mathbf{p}_0\in\mathbb{R}^{d_{\mathrm{lat}}}$.
The map induces a conditional distribution over latent initial states, enabling diverse trajectory generation for a given system.
We choose $d_{\mathrm{lat}}$ as an upper bound on the expected effective dimensionality of the observed dynamics (see Appendix~\ref{subsec:ablation_phasespace_dimension}
for sensitivity analysis).
To encourage a compact representation within this potentially over-parameterized space, we introduce cyclic-coordinate regularization in Section~\ref{subsec:latent_space_motion_model}.
\subsection{Phase-Space Dynamics}\label{subsec:latent_space_motion_model}
Starting from the initial state $\mathbf{z}_0$ produced by the phase-space map, SPS-GAN generates a latent trajectory $\mathbf{z}_{0:T}=[\mathbf{z}_0,\ldots,\mathbf{z}_T]$ using a pHNN conditioned on system information $\xi$,
\begin{equation}
    \mathbf{z}_{t+1} = \mathbf{z}_{t} + \int_{t}^{t+1}\left(\left(\begin{bmatrix}
        0 & \mathbf{I}\\ -\mathbf{I} & 0 \end{bmatrix} - \begin{bmatrix} 0 & 0 \\ 0 & \mathbf{D}_{\phi}\left(\xi\right) \end{bmatrix}\right)\nabla_{\mathbf{z}}\mathcal{H}_{\theta}(\mathbf{z}_{\tau},\xi) + \begin{bmatrix} 0 \\ \mathbf{F}_{\psi}\left(\mathbf{u}_{\tau},\xi\right)\end{bmatrix}\right)d\tau.\label{eqn:sps_phnn}
\end{equation}
Here, $\mathcal{H}_{\theta}(\mathbf{z},\xi)$ is the learned Hamiltonian, while $\mathbf{D}_{\phi}(\xi)$ and $\mathbf{F}_{\psi}(\mathbf{u},\xi)$ model dissipation and external forcing, respectively.
The dissipation matrix is constrained to be positive semi-definite, as described in Appendix~\ref{subsec:result_traj_rollout}.
For conservative systems, setting the non-conservative terms to zero recovers standard Hamiltonian neural dynamics.

When the latent phase space is over-parameterized, learned dynamics may occupy more coordinates than are needed to represent the observed motion.
To encourage compact latent representations, we adapt the cyclic-coordinate regularization introduced in~\cite{allen2024hamiltonian}.
In conservative Hamiltonian systems, a cyclic coordinate has a conserved conjugate momentum.
In port-Hamiltonian systems, however, dissipation and forcing can change the conjugate momentum even when the corresponding coordinate is cyclic.
From Equation~\eqref{eqn:sps_phnn}, the momentum dynamics are
\begin{equation}
    \dot{\mathbf{p}} = -\nabla_{\mathbf{q}}\mathcal{H}_{\theta}\left(\mathbf{z},\xi\right) - \mathbf{D}_{\phi}\left(\xi\right)\nabla_{\mathbf{p}}\mathcal{H}_{\theta}\left(\mathbf{z},\xi\right) + \mathbf{F}_{\psi}\left(\mathbf{u},\xi\right).
    \label{eqn:sps_momentum}
\end{equation}
To isolate the Hamiltonian contribution from dissipation and forcing, we define the cyclic-coordinate loss
\begin{equation}
    \mathcal{L}_{\mathrm{cyclic}} = \lambda_{\mathrm{cyclic}}\sum\nolimits_{i=1}^{d_{\mathrm{lat}}}\left|\dot{p}_{i} + \left[\mathbf{D}_{\phi}(\xi)\nabla_{\mathbf{p}}\mathcal{H}_{\theta}(\mathbf{z},\xi) - \mathbf{F}_{\psi}\left(\mathbf{u}, \xi\right)\right]_{i}\right|.
    \label{eqn:cyclic_coord_loss}
\end{equation}
By Equation~\eqref{eqn:sps_momentum}, this loss is equivalent to penalizing the coordinate derivatives of the Hamiltonian, $\lambda_{\mathrm{cyclic}}\sum_i |\partial\mathcal{H}_{\theta}/\partial q_i|$.
It therefore encourages the learned Hamiltonian to become independent of selected coordinates without penalizing momentum changes arising from non-conservative effects.
Following~\cite{desai2021port}, we additionally regularize the dissipation and forcing terms,
\begin{equation}
    \mathcal{L}_{\mathrm{non\text{-}cons}} = \lambda_{D}\lVert\mathbf{D}_{\phi}(\xi)\rVert + \lambda_{F}\lVert\mathbf{F}_{\psi}(\mathbf{u}, \xi)\rVert. \label{eqn:noncon_reg}
\end{equation}
This penalty discourages unnecessary non-conservative contributions, while cyclic-coordinate regularization encourages compact latent mechanical representations.
Cyclic-coordinate regularization is intended to encourage compact latent representations. We evaluate the resulting active dimensionality and mechanical interpretability in Sections~\ref{subsec:result_DoF} and~\ref{subsec:results_interpretability}.
\subsection{Generating Cartesian Trajectories}\label{Sec:Method_Traj}
To generate Cartesian trajectories, SPS-GAN decodes each latent phase-space state using an MLP generator $G_{\mathrm{T}}$,
\begin{equation}
    \tilde{\mathbf{x}}_{t} = G_{\mathrm{T}}\left(\mathbf{z}_{t}\right) \in \mathbb{R}^{ d_{\mathrm{out}}},
\end{equation}
where $d_{\mathrm{out}}$ is the maximum flattened Cartesian observation dimension across the systems in the dataset.
To accommodate systems with different numbers of particles, we apply a system-specific binary mask $\mathbf{M}(\xi)$,
\begin{equation}
    \hat{\mathbf{x}}_{t} = \tilde{\mathbf{x}}_{t} \odot \mathbf{M}(\xi),
\end{equation}
which removes unused output coordinates.
The generated trajectory is $\hat{\mathbf{x}}_{0:T} =[\hat{\mathbf{x}}_{0},\ldots,\hat{\mathbf{x}}_{T}]$, and real trajectories are padded to the same dimension.

We train a recurrent discriminator $D_{\mathrm{T}}\left(\cdot,\xi\right))$ to distinguish generated trajectory subsequences from real observations, conditioned on system information $\xi$.
The conditional adversarial objective is
\begin{equation}
    V_{\mathrm{traj}}(G_{\mathrm{T}},D_{\mathrm{T}}) = \mathbb{E}_{\mathbf{x}\sim p_{X}(\cdot| \xi)}\left[\log D_{\mathrm{T}}\left(\mathbf{x}_{s:s+\ell},\xi\right)\right] + \mathbb{E}_{\hat{\mathbf{x}}\sim p_{G_{\mathrm{T}}(\cdot| \xi)}}\left[\log \left(1-D_{\mathrm{T}}\big(\hat{\mathbf{x}}_{s:s+\ell},\xi\big)\right)\right].
\end{equation}
The generator, phase-space map, and pHNN are trained jointly using the adversarial objective and the cyclic-coordinate and non-conservative regularization terms introduced in Section~\ref{subsec:latent_space_motion_model}.
We evaluate the dynamical consistency of the generated Cartesian trajectories against supervised prediction baselines in Section~\ref{sec:results_traj}.
\subsection{Generating Videos}\label{Sec:Method_Video}
SPS-GAN uses the same phase-space map, latent dynamics, and cyclic-coordinate regularization across observation modalities, adapting only the observation generator and discriminators.
For video observations, a convolutional generator $G_{\mathrm{I}}$ decodes each latent state $\mathbf{z}_t$ into an RGB frame.
To model appearance variation separately from motion, we sample a content code $\boldsymbol{\epsilon}_{\mathrm{c}}\sim\mathcal{N}(0,\mathbf{I}_{d_{\mathrm{c}}})$ and hold it fixed across the generated sequence.
The generator produces
\begin{equation}
    \hat{\mathbf{v}}_{t} = G_{\mathrm{I}}\left([\mathbf{z}_{t}, \boldsymbol{\epsilon}_{\mathrm{c}}]\right)
    \in \mathbb{R}^{d_{\mathrm{h}}\times d_{\mathrm{w}}\times 3},
\end{equation}
yielding the video sequence $\hat{\mathbf{v}}_{0:T} = [\hat{\mathbf{v}}_{0}, \ldots, \hat{\mathbf{v}}_{T}]$.

To encourage both visual realism and temporal coherence, we use two discriminators conditioned on system information $\xi$.
The image discriminator $D_{\mathrm{I}}\left(\cdot,\xi\right)$ evaluates individual frames, while the video discriminator $D_{\mathrm{V}}\left(\cdot,\xi\right)$ evaluates sampled subsequences $\mathbf{v}_{s:s+\ell}$.
The corresponding adversarial objective is
\begin{align}
\begin{split}
    V_{\mathrm{video}}(G_{\mathrm{I}},&D_{\mathrm{I}},D_{\mathrm{V}}) = \mathbb{E}_{\mathbf{v}\sim p_{V}(\cdot| \xi)}\left[\log D_{\mathrm{V}}(\mathbf{v}_{s:s+\ell},\xi) + \frac{1}{T}\sum\nolimits_{t=0}^{T}\log D_{\mathrm{I}}(\mathbf{v}_{t},\xi)\right] \\
    & +\mathbb{E}_{\hat{\mathbf{v}}\sim p_{G_{\mathrm{I}}}(\cdot| \xi)}\left[\log\left(1-D_{\mathrm{V}}(\hat{\mathbf{v}}_{s:s+\ell},\xi)\right) + \frac{1}{T}\sum\nolimits_{t=0}^{T}\log\left(1-D_{\mathrm{I}}(\hat{\mathbf{v}}_{t},\xi)\right)\right].
\end{split}
\end{align}
The video generator, phase-space map, and pHNN are trained jointly using the corresponding generator loss and the regularization terms $\mathcal{L}_{\mathrm{cyclic}}$ and $\mathcal{L}_{\mathrm{non\text{-}cons}}$.
We evaluate the visual fidelity and temporal consistency of the generated sequences in Section~\ref{sec:results_video}.
\section{Results}\label{sec:results}
We evaluate SPS-GAN along two dimensions, focusing on the dynamical fidelity of generated motion and the dimensionality and interpretability of the learned latent phase space.
Experiments span ordered, chaotic, dissipative, and forced systems using simulated and real-world Cartesian trajectories and video observations.
We assess dynamical fidelity through initial-condition-aligned comparisons with supervised trajectory prediction models, and evaluate video quality and temporal consistency against generative baselines.
We further examine whether the learned representations exhibit the expected effective dimensionality and whether their active directions correspond to interpretable mechanical coordinates.
\subsection{Generating Cartesian Trajectories}
\label{sec:results_traj}
\begin{table}[b]
    \vspace{-0.15in}
    \caption{\textbf{Trajectory generation accuracy on conservative systems.} We report MSE between each generated or predicted trajectory and the simulator rollout initialized from the same state. SPS-GAN is trained per system, while cSPS-GAN is trained jointly across all five systems with system labels. Best results are bolded and second-best results are underlined.}
    \label{tab:result_conservative_accuracy}
    \centering
    \vskip 0.03in
    \renewcommand{\arraystretch}{0.7}
    \begin{tabular}{lccccc}
    \toprule
    Network & Mass Spring & Single Pend. & Double Pend. & Two Body & Three Body\\
    \midrule
    SPS-GAN & $6.16\times10^{-4}$ & $\boldsymbol{4.29\times10^{-4}}$ & $\boldsymbol{3.25\times10^{-2}}$ & $\boldsymbol{3.07\times10^{-3}}$ & $\boldsymbol{1.26\times10^{-3}}$\\
    HNN & $3.53\times10^{-4}$ & $6.27\times10^{-4}$ & $1.25\times10^{-1}$ &  $\underline{\smash{3.61\times10^{-3}}}$ & $2.81\times10^{-3}$\\
    SymODEN & $6.40\times10^{-4}$ & $6.56\times10^{-4}$ & $9.89\times10^{-2}$ & $6.20\times10^{-3}$ & $8.08\times10^{-3}$\\
    TRS-ODEN & $6.72\times10^{-4}$ & $9.57\times10^{-4}$ & $6.12\times10^{-2}$ & $6.36\times10^{-3}$ & $6.22\times10^{-3}$\\
    NeuralODE & $\underline{\smash{1.62\times10^{-4}}}$ & $1.59\times10^{-3}$ & $5.99\times10^{-2}$ & $5.72\times10^{-3}$ & $3.67\times10^{-3}$\\
    LatentODE & $\boldsymbol{1.13\times10^{-4}}$ & $9.27\times10^{-4}$ & $4.66\times10^{-2}$ & $1.98\times10^{-2}$ & $\underline{\smash{1.37\times10^{-3}}}$\\
    HGN & $1.81\times10^{-3}$ & $1.92\times10^{-2}$ & $2.27\times10^{-1}$ & $9.31\times10^{-1}$ & $7.99\times10^{-2}$\\
    TimeGAN & $1.17\times10^{-2}$ & $2.43\times10^{-2}$ & $5.82\times10^{-1}$ & $1.21\times10^{0}$ & $1.97\times10^{0}$\\
    Koopman VAE & $6.84\times10^{-4}$ & $5.27\times10^{-3}$ & $5.46\times10^{-2}$ & $9.36\times10^{-3}$ & $1.13\times10^{-2}$\\
    DiffusionTS & $7.54\times10^{-4}$ & $5.52\times10^{-3}$ & $5.90\times10^{-2}$ & $3.83\times10^{-3}$ & $1.22\times10^{-2}$\\
    \midrule
    cSPS-GAN & $9.88\times10^{-4}$ & $\underline{\smash{5.11\times10^{-4}}}$ & $\underline{\smash{3.33\times10^{-2}}}$ & $4.46\times10^{-3}$ & $2.61\times10^{-3}$\\
    \bottomrule
    \end{tabular}
\end{table}
We first evaluate the dynamical consistency of trajectories generated by SPS-GAN.
Unlike predictive baselines that evolve prescribed initial conditions, SPS-GAN samples trajectories from a learned distribution, requiring an initial-condition alignment protocol for comparison.
For each generated trajectory $\hat{\mathbf{x}}_{0:T}$, we initialize the true simulator and predictive baselines at $\hat{\mathbf{x}}_0$, then compute the MSE of each rollout against the simulator trajectory.
This evaluates whether SPS-GAN generates trajectories consistent with the underlying dynamics rather than reconstructing particular observed trajectories.

We evaluate seven systems, namely mass-spring, single pendulum, double pendulum, two-body, three-body, damped pendulum, and forced Duffing, which spans conservative, dissipative, forced, and chaotic dynamics.
Details on training data and example rollouts are provided in  Appendices~\ref{apn:dataset} and \ref{subsec:result_traj_rollout}.

\textbf{Conservative Systems.}
We evaluate whether SPS-GAN can accurately capture conservative dynamics under both single-system and joint conditional training.
We consider two configurations: SPS-GAN, trained independently on each system, and cSPS-GAN, a single conditional model trained jointly across all five systems using system labels.
For both configurations, we set $\mathbf{D}_{\phi}=\mathbf{F}_{\psi}=0$.
We compare against supervised dynamics baselines HNN~\citep{greydanus2019hamiltonian}, SymODEN~\citep{zhong2019symplectic}, NeuralODE~\citep{chen2018neural}, LatentODE~\citep{rubanova2019latent}, and TRS-ODEN~\citep{huh2020time}, and generative baselines HGN~\citep{toth2019hamiltonian}, TimeGAN~\citep{yoon2019time}, Koopman VAE~\citep{naiman2024generative}, and DiffusionTS~\citep{yuan2024diffusion}.
As shown in Table~\ref{tab:result_conservative_accuracy}, SPS-GAN achieves the lowest MSE on four of the five systems.
cSPS-GAN achieves comparable accuracy on most systems while modeling all five systems jointly.
These results indicate that SPS-GAN can generate dynamically consistent trajectories without prescribed initial conditions, and that conditional training enables multiple systems to share a single model.

\textbf{Long-Horizon Distributional Fidelity.}
In chaotic systems, pointwise trajectory errors can grow rapidly and do not necessarily reflect agreement in long-term motion statistics~\citep{brenner2022tractable,wu2026f3no}.
We therefore evaluate the long-horizon distributional fidelity of SPS-GAN on the chaotic double pendulum.
We train all models with a rollout horizon of $T=200$ and compare 4,096 generated rollouts with 4,096 simulated reference rollouts.
We compare the distributions of visited phase-space states using sliced Wasserstein distance (SWD), Earth mover's distance (EMD), and forward Kullback--Leibler (KL) divergence.
We further assess energy-distribution agreement, energy conservation, and recurrence statistics using energy SWD, per-rollout energy drift, and Poincar\'{e}-section SWD, respectively.
As shown in Table~\ref{tab:result_distributional_fidelity}, SPS-GAN achieves the lowest values across all six metrics among the evaluated baselines.
These results indicate that SPS-GAN captures complementary aspects of the system's long-term behavior, including phase-space occupancy, energy statistics, and recurrence structure.

\textbf{Non-Conservative Systems.} 
We next evaluate whether the port-Hamiltonian formulation supports accurate generation of dissipative and externally forced dynamics.
We evaluate SPS-GAN on a damped pendulum and forced Duffing oscillator, with the learned dissipation and forcing components of the model enabled.
As shown in Table~\ref{tab:result_nonconservative_accuracy}, SPS-GAN achieves the lowest MSE on both systems when compared against pHNN, NeuralODE, LatentODE, and the dissipative variant of TRS-ODEN.
We additionally evaluate an ideal pendulum as a conservative control, where SPS-GAN also achieves the lowest MSE with the full port-Hamiltonian parameterization.
\begin{table}[t]
    \caption{\textbf{Long-horizon distributional fidelity on the chaotic double pendulum.} We compare generative and deterministic models using phase-space distribution distances, energy-distribution agreement, relative energy drift, and Poincar\'e-section recurrence statistics. Lower values indicate better performance.}
    \label{tab:result_distributional_fidelity}
    \centering
    \vskip 0.03in
    \renewcommand{\arraystretch}{0.7}
    \begin{tabular}{lccccccc}
    \toprule
    \multirow{2}{*}{Network} & \multirow{2}{*}{Type} & \multicolumn{3}{c}{Phase Space} & \multicolumn{2}{c}{Energy} & Poincar\'e\\
    \cmidrule(lr){3-5} \cmidrule(lr){6-7} \cmidrule(lr){8-8}
    & & SWD & EMD & KL & SWD & Drift & SWD\\
    \midrule
    SPS-GAN & Gen. & $\boldsymbol{0.054}$ & $\boldsymbol{0.256}$ & $\boldsymbol{0.370}$ & $\boldsymbol{0.056}$ & $\boldsymbol{4.521\%}$ & $\boldsymbol{0.661}$\\
    TimeGAN & Gen. & $0.126$ & $0.397$ & $1.146$ & $0.287$ & $25.113\%$ & $1.588$\\
    Koopman VAE & Gen. & $0.281$ & $0.584$ & $0.713$ & $0.501$ & $15.751\%$ & $1.853$\\
    Diff-TS & Gen. & $\underline{\smash{0.090}}$ & $\underline{\smash{0.305}}$ & $\underline{\smash{0.554}}$ & $0.472$ & $29.203\%$ & $1.643$\\
    HGN & Gen. & $0.147$ & $0.349$ & $0.679$ & $0.206$ & $9.391\%$ & $0.979$\\
    \midrule
    NeuralODE & Det. & $0.329$ & $0.667$ & $0.994$ & $0.507$ & $24.721\%$ & $1.327$\\
    LatentODE & Det. & $0.241$ & $0.544$ & $0.561$ & $0.357$ & $23.608\%$ & $1.134$\\
    SymODEN & Det. & $0.212$ & $0.707$ & $2.563$ & $\underline{\smash{0.144}}$ & $\underline{\smash{7.159\%}}$ & $0.938$\\
    HNN & Det. & $0.393$ & $0.992$ & $2.504$ & $0.181$ & $11.611\%$ & $\underline{\smash{0.785}}$\\
    \bottomrule
    \end{tabular}
    \vspace{-0.15in}
\end{table}
\begin{table}[b]
    \vspace{-0.15in}
    \caption{\textbf{Trajectory generation accuracy on non-conservative systems.} We report MSE under the same initial-condition alignment protocol as Table~\ref{tab:result_conservative_accuracy}. SPS-GAN uses the port-Hamiltonian formulation for dissipative and forced systems; the ideal pendulum is used as a conservative control.}
    \label{tab:result_nonconservative_accuracy}
    \centering
    \vskip 0.03in
    \renewcommand{\arraystretch}{0.7}
    \begin{tabular}{lccc}
    \toprule
    Network & Ideal Pend. & Damped Pend. & Forced Duffing\\
    \midrule
    SPS-GAN & $\boldsymbol{4.77\times10^{-4}}$ & $\boldsymbol{4.57\times10^{-4}}$ & $\boldsymbol{5.22\times10^{-4}}$\\
    pHNN & $\underline{\smash{8.42\times10^{-4}}}$ & $\underline{\smash{4.76\times10^{-4}}}$ & $	6.92\times10^{-4}$\\
    NeuralODE & $1.59\times10^{-3}$ & $6.85\times10^{-4}$ & $7.35\times10^{-4}$\\
    LatentODE & $9.27\times10^{-4}$ & $6.69\times10^{-4}$ & $1.18\times10^{-3}$\\
    TRS-ODEN & $9.57\times10^{-4}$ & $1.44\times10^{-3}$ & $\underline{\smash{6.50\times10^{-4}}}$\\
    \bottomrule
    \end{tabular}
\end{table}
These results indicate that the framework accommodates non-conservative dynamics while retaining accuracy when dissipation and forcing are unnecessary.
Additional experiments on an actuated pendulum with external control inputs are provided in Appendix~\ref{subsec:result_control_pend}.
\subsection{Generating Video Sequences}
\label{sec:results_video}
We evaluate whether SPS-GAN generates visually realistic videos with temporally consistent, physically plausible motion.
Experiments consider rendered conservative systems, conditional generation under physical and appearance variation, and real-world single- and double-pendulum videos.
Details on data generation and training are provided in Appendix~\ref{apn:dataset}, with example rollouts in Appendix~\ref{subsec:result_video_rollout}.

\begin{table}[t]
    \caption{\textbf{Video generation quality and physical consistency.} We report FVD against baselines and energy variance ($\Delta E$) of centroid-extracted trajectories. SPS-GAN achieves the lowest FVD on all rendered conservative systems, while its $\Delta E$ closely matches the simulator reference floor.}
    \label{tab:result_single_conservative_video}
    \centering
    \vskip 0.03in
    \renewcommand{\arraystretch}{0.7}
    \begin{tabular}{llccccc}
        \toprule
        & \multirow{2}{*}{Network} & Mass & Single & Double & Two & Three\\
        & & Spring & Pend. & Pend. & Body & Body\\
        \midrule
        \multirow{3}{*}{FVD} 
        & SPS-GAN & \textbf{25.6$\pm$3.8} & \textbf{40.6$\pm$6.9} & \textbf{24.1$\pm$3.8} & \textbf{87.1$\pm$7.9} & \textbf{89.1$\pm$11.4}\\
        & HGAN & \underline{\smash{45.7$\pm$3.8}} & \underline{\smash{91.6$\pm$8.2}} & \underline{\smash{73.2$\pm$7.1}} & \underline{\smash{105.9$\pm$8.4}} & 1981.1$\pm$98.6\\
        & HGN & 385.1$\pm$26.8 & 688.1$\pm$85.6 & 331.9$\pm$41.2 & 830.9$\pm$72.3 & \underline{\smash{451.4$\pm$60.3}}\\
        \midrule
        \multirow{2}{*}{$\Delta E$} 
        & SPS-GAN & $2.33\%$ & $2.89\%$ & $6.83\%$ & $5.96\%$ & $3.47\%$\\
        & Simulator & $1.64\%$ & $2.62\%$ & $6.32\%$ & $6.11\%$ & $3.57\%$\\
        \bottomrule
    \end{tabular}
    \vspace{-0.15in}
\end{table}
\textbf{Fixed-Parameter Conservative Systems.} 
We first evaluate the visual fidelity and physical consistency of videos generated from conservative systems.
SPS-GAN is trained separately on rendered observations of five conservative systems, without access to their underlying physical states.
As shown in Table~\ref{tab:result_single_conservative_video}, SPS-GAN achieves the lowest Fréchet Video Distance (FVD)~\cite{unterthiner2019fvd} on all five systems, outperforming generative baselines HGN~\citep{toth2019hamiltonian} and HGAN~\citep{allen2024hamiltonian}.
Because FVD alone does not establish physical consistency, we extract particle centroids from generated videos and measure the energy variance along each rollout.
We apply the same procedure to simulator videos, whose nonzero energy variance reflects errors introduced by centroid extraction rather than physical energy drift.
SPS-GAN produces energy variation comparable to this simulator reference across all five systems, providing evidence that its generated videos approximately preserve the expected conservative behavior.

\textbf{Variable-Parameter Conservative Systems.} 
We next evaluate whether a single conditional SPS-GAN model can generate videos across multiple systems while accommodating variation in physical and visual parameters.
We train cSPS-GAN jointly across systems using conditional labels $\xi$ specifying system identity and parameter settings, including particle size and color.
Figure~\ref{fig:CHGAN_TMLR_VVVComparison} in Appendix~\ref{subsec:result_video_rollout} illustrates generated rollouts under different physical and visual conditions. 
cSPS-GAN achieves FVDs of $135.3\text{--}182.2$, outperforming the single-system HGN results reported in Table~\ref{tab:result_single_conservative_video}.
These results demonstrate that a single conditional model can generate videos across multiple systems and parameter configurations.

\begin{figure}[b]
    \vspace{-0.12in}
    \centering
    \includegraphics[width=\textwidth]{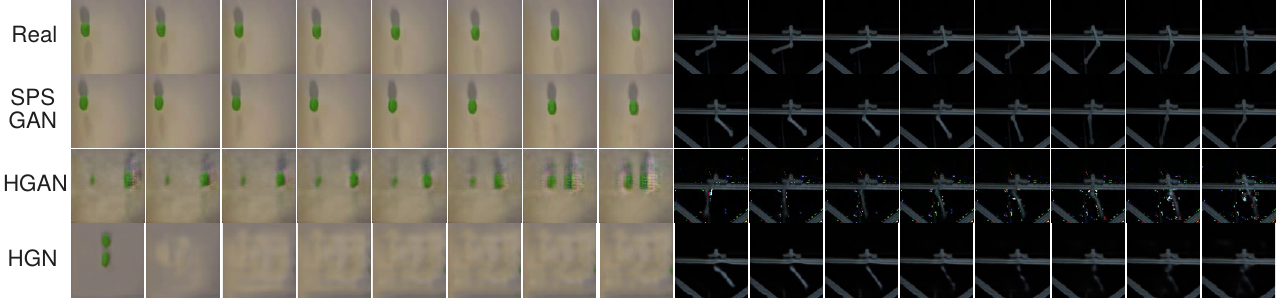}
    \vskip -0.06in
    \caption{\textbf{Video generation on real-world pendulum systems.} Qualitative comparison of SPS-GAN, HGN, and HGAN on \textbf{(left)} single-pendulum~\citep{garcia2025learning} and \textbf{(right)} double-pendulum~\citep{kaheman2023experimental} videos. SPS-GAN generates temporally coherent motion, whereas the baselines exhibit static or dynamically inconsistent sequences. For the single pendulum, the relative energy variance of centroid-extracted SPS-GAN videos ($2.9\%$) is close to that of real recordings ($2.7\%$).}
    \label{fig:SPSGAN_RealDPVideoComparison}
\end{figure}

\textbf{Real-World Systems.} We next evaluate whether SPS-GAN generates dynamically consistent videos from real-world observations, where non-conservative effects and complex visual appearances depart from idealized simulations.
We train SPS-GAN on real-world videos of a freely swinging single pendulum~\citep{garcia2025learning} and double pendulum~\citep{kaheman2023experimental}.
Figure~\ref{fig:SPSGAN_RealDPVideoComparison} shows that SPS-GAN generates temporally coherent pendulum motion, whereas HGN produces nearly static sequences and HGAN exhibits visibly inconsistent motion.
For the single pendulum, the energy variance of the centroid-extracted SPS-GAN videos is $2.9\%$, closely matching the $2.7\%$ measured from real videos.
These results suggest that SPS-GAN can generate coherent motion from real-world video observations while reproducing aspects of the underlying energy behavior.
\subsection{Identifying Active Degrees of Freedom}\label{subsec:result_DoF}
\begin{table}[t]

    \caption{\textbf{Identifying active degrees of freedom.} We report the percentage variance explained (PVE) by the leading principal components of the learned phase space. Components contributing to 95\% cumulative PVE are highlighted in {\setlength{\fboxsep}{1pt}\colorbox{LimeGreen!35}{green}}. The estimated active dimensionality matches the expected effective degrees of freedom across simulated and real-world trajectory and video datasets.}
    \label{tab:result_DoF}
    \centering
    \vskip 0.03in
    \setlength{\tabcolsep}{3.8pt}
    \renewcommand{\arraystretch}{0.7}
    \begin{tabular}{lccccccccccc}
    \toprule
    & Mass & Sin. & Dou. & Two & Three & Damp. & Force. & Sin. & Dou. & Sin. & Dou.\\
    & Spring & Pend. & Pend. & Body & Body & Pend. & Duff. & Pend. & Pend. & Pend. & Pend.\\
    \midrule
    Mode & Traj. & Traj. & Traj. & Traj. & Traj. & Traj. & Traj. & Traj. & Traj. & Vid. & Vid.\\
    \midrule
    Source & Sim & Sim & Sim & Sim & Sim & Sim & Sim & Real & Real & Real & Real\\
    \midrule
    $\mathrm{PVE}_{1}$ & {\setlength{\fboxsep}{1pt}\colorbox{LimeGreen!35}{98.85}} & {\setlength{\fboxsep}{1pt}\colorbox{LimeGreen!35}{97.16}} & {\setlength{\fboxsep}{1pt}\colorbox{LimeGreen!35}{78.70}} & {\setlength{\fboxsep}{1pt}\colorbox{LimeGreen!35}{95.87}} & {\setlength{\fboxsep}{1pt}\colorbox{LimeGreen!35}{81.87}} & {\setlength{\fboxsep}{1pt}\colorbox{LimeGreen!35}{95.79}} & {\setlength{\fboxsep}{1pt}\colorbox{LimeGreen!35}{96.47}} & {\setlength{\fboxsep}{1pt}\colorbox{LimeGreen!35}{96.28}} & {\setlength{\fboxsep}{1pt}\colorbox{LimeGreen!35}{61.39}} & {\setlength{\fboxsep}{1pt}\colorbox{LimeGreen!35}{98.69}} & {\setlength{\fboxsep}{1pt}\colorbox{LimeGreen!35}{70.47}}\\
    $\mathrm{PVE}_{2}$ & 0.48 & 0.79 & {\setlength{\fboxsep}{1pt}\colorbox{LimeGreen!35}{18.39}} & 1.32 & {\setlength{\fboxsep}{1pt}\colorbox{LimeGreen!35}{17.40}} & 1.28 & 1.02 & 1.91 & {\setlength{\fboxsep}{1pt}\colorbox{LimeGreen!35}{35.41}} & 0.64 & {\setlength{\fboxsep}{1pt}\colorbox{LimeGreen!35}{24.63}}\\
    $\mathrm{PVE}_{3}$ & 0.18 & 0.58 & 0.90 & 1.05 & 0.24 & 0.79 & 0.76 & 0.70 & 1.96 & 0.46 & 0.88\\
    \midrule
    DoF & 1 & 1 & 2 & 1 & 2 & 1 & 1 & 1 & 2 & 1 & 2\\
    \bottomrule
    \end{tabular}
    \vspace{-0.15in}
\end{table}
We next examine whether the learned latent phase space captures the effective degrees of freedom of the observed dynamics.
The effective degrees of freedom characterize the number of independent coordinates needed to describe the observed motion.
Identifying this dimensionality from trajectory or video observations provides insight into the underlying mechanical structure without requiring prescribed generalized coordinates.
SPS-GAN encourages compact latent representations through cyclic-coordinate regularization, and we evaluate their active dimensionality using principal component analysis (PCA).
Specifically, we perform PCA~\citep{abdi2010principal,Jolliffe20114PCA} on the learned coordinate trajectories $\mathbf{q}_{0:T}$ from 4,096 randomly sampled phase-space rollouts.
We compute the percentage variance explained (PVE) by principal component $k$ as
\begin{equation}
    \mathrm{PVE}_{k} = \nicefrac{\lambda_{k}}{\sum_{i} \lambda_{i}},
\end{equation}
where $\lambda_k$ is the corresponding eigenvalue of the covariance matrix.
We estimate the active dimension as the smallest number of components whose cumulative PVE exceeds 95\%~\citep{james2021introduction}.
Full details and 10-component PVE results are provided in Appendix~\ref{apn:results_dof}.

We use $d_{\mathrm{lat}}=10$ across all experiments, providing an overparameterized latent phase space relative to the expected effective degrees of freedom.
We evaluate five conservative and two non-conservative simulated systems, two real-world Cartesian trajectory datasets~\citep{kaheman2023experimental,schmidt2009distilling}, and two real-world video datasets~\citep{garcia2025learning,kaheman2023experimental}.
As shown in Table~\ref{tab:result_DoF}, the number of principal components required to explain 95\% of the variance matches the expected effective degrees of freedom in all 11 settings.
The learned representations exhibit one active dimension for the mass-spring, single-pendulum, and two-body systems, and two for the double-pendulum and three-body systems.
The expected dimensions are also identified from real-world trajectory and video observations, suggesting that these findings extend beyond idealized simulated data.
Together, these results indicate that the dominant directions of variation in SPS-GAN's learned latent coordinates reflect the effective dimensionality of the observed dynamics.
Analytical derivations of the effective degrees of freedom are provided in Appendix~\ref{subsec:DoF_canonical_coordinates}.
\subsection{Identifying Interpretable Mechanical Coordinates}
\label{subsec:results_interpretability}
\begin{figure}[b]
    \vspace{-0.15in}
    \centering
    \includegraphics[width=\textwidth]{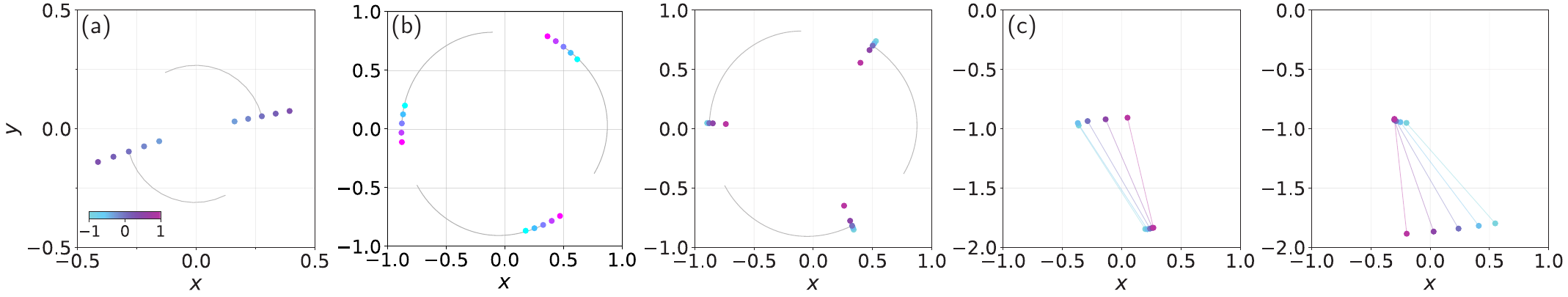}
    \vskip -0.05in
    \caption{\textbf{Interpretable coordinates from active latent directions.} Sweeping active latent directions and decoding to Cartesian space reveals coordinates aligned with known mechanical variables: \textbf{(a)} radial distance for two-body; \textbf{(b)} orbital angle and radial scale for three-body; \textbf{(c)} two angular positions for double pendulum.}
    \label{fig:SPSGAN_CanonicalCoordinates}
\end{figure}
Identifying the active latent dimensions establishes the effective dimensionality of the learned representation, but does not reveal the physical quantities those dimensions encode.
We therefore examine whether the active latent directions correspond to known mechanical coordinates by varying them individually and decoding the resulting states into observation space.

We use the PCA decomposition from Section~\ref{subsec:result_DoF} to identify the active principal directions $\mathbf{a}_k$ of the learned generalized coordinates.
To interpret each direction, we vary its coordinate while holding the remaining principal components and latent momenta fixed.
Specifically, we sample five states along each direction and decode them into Cartesian space,
\begin{equation}
    \mathbf{q}_{k}^{(i)} = \mathbf{q}^{(0)} + i\sqrt{\lambda_{k}}\mathbf{a}_{k},
    \qquad
    \mathbf{z}_{k}^{(i)} = (\mathbf{q}_{k}^{(i)}, \mathbf{p}^{(0)}),
    \qquad
    \mathbf{x}_{k}^{(i)} = G_{\mathrm{T}}\left(\mathbf{z}_{k}^{(i)}\right),
\end{equation}
$\lambda_k$ is the corresponding PCA eigenvalue, and $(\mathbf{q}^{(0)},\mathbf{p}^{(0)})$ is a reference latent state.
The decoded sweeps allow us to examine how each active latent direction relates to physical motion.

Figure~\ref{fig:SPSGAN_CanonicalCoordinates} shows decoded sweeps for three representative systems.
For the two-body system, the active direction corresponds to changes in radial separation $r$.
For the three-body system, the two active directions correspond to radial scale and orbital angle, while those of the double pendulum correspond to the two pendulum angles $\theta_1$ and $\theta_2$.
Additional results for the mass-spring, single pendulum, and single-pendulum video are provided in Appendix~\ref{apn:results_cancor}.
These results demonstrate that the learned active directions can correspond to physically interpretable mechanical coordinates, despite training only on Cartesian trajectories or video observations.
%
\begin{table}[t]
    \caption{\textbf{Ablation analysis on two-body system.} We ablate the phase-space map, cyclic-coordinate loss, and pHNN module. Each ablation degrades both accuracy and recovered active dimension.}
    \label{tab:ablation}
    \centering
    \vskip 0.03in
    \setlength{\tabcolsep}{5pt}
    \renewcommand{\arraystretch}{0.85}
    \begin{tabular}{lccccc}
    \toprule
    & SPS-GAN & +SympNet & +GRU & w/o Phase Space Map & w/o Cyclic Loss\\
    \midrule
    MSE & $\boldsymbol{3.07\times10^{-3}}$ & $4.43\times10^{-3}$ & $1.60\times10^{-1}$ & $4.93\times10^{-2}$ & $4.40\times10^{-2}$\\
    DoF & 1 & 2 & 3 & 7 & 3\\
    \bottomrule
    \end{tabular}
    \vspace{-0.15in}
\end{table}
\subsection{Ablation Analysis}\label{subsec:results_ablation}
We evaluate the contributions of the phase-space map, cyclic-coordinate regularization, and structured latent dynamics to trajectory accuracy and active dimensionality.
On the two-body system, we remove the phase-space map or cyclic-coordinate loss, and replace the pHNN dynamics with either SympNet~\citep{jin2020sympnets} or GRU~\citep{cho2014learning}.
As shown in Table~\ref{tab:ablation}, the complete SPS-GAN achieves the lowest trajectory MSE and recovers the expected single active dimension.
Replacing the pHNN with SympNet or GRU increases trajectory error and the estimated active dimension, with the GRU exhibiting substantially larger errors.
Removing the phase-space map produces the largest dimensionality discrepancy, increasing the estimated active dimension from one to seven, while removing cyclic-coordinate regularization increases it to three.
These results indicate that the initialization map, structured dynamics, and cyclic-coordinate regularization make complementary contributions to dynamical fidelity and compact latent representations.

Additional ablations and robustness analyses are provided in Appendix~\ref{apn:ablation_robustness}.
These examine sensitivity to latent dimensionality, PCA thresholds, observation noise, training progress, and regularization weights.
We also evaluate dimensionality estimates using eigenvalue-ratio and inactive-component pruning diagnostics, and assess the coverage of generated initial conditions.
\section{Conclusion}
We introduced SPS-GAN, a generative framework that combines structured latent dynamics with adversarial learning to generate trajectories and videos while discovering interpretable mechanical structure.
By integrating port-Hamiltonian dynamics with cyclic-coordinate regularization, SPS-GAN models conservative, dissipative, and forced systems across trajectory and video observations.
Experiments on simulated and real-world systems demonstrate strong dynamical fidelity and video-generation quality relative to established baselines.
The results suggest that generative models equipped with structured latent dynamics can discover compact, physically interpretable representations directly from trajectory and video observations.

\textbf{Limitations.}
SPS-GAN assumes sufficiently smooth dynamics that admit an approximate port-Hamiltonian representation, limiting its applicability to strongly stochastic or discontinuous systems.
Active dimensionality is estimated from a finite-sample PCA spectrum and is sensitive to the variance threshold and observation noise.
While dimensionality estimates remain consistent for thresholds of 90-95\% and signal-to-noise ratios down to 10~dB in our experiments, stricter thresholds or greater noise can lead to overestimation.
Finally, real-world evaluation is limited to low-dimensional trajectory recordings and planar video observations, leaving fully three-dimensional and contact-rich dynamics for future investigation.
\begin{ack}
Yulong Yang was supported by a grant from the National Aeronautics and Space Administration funded through the Jet Propulsion Laboratory Strategic University Research Partnerships program. 
We sincerely thank the NeurIPS 2026 reviewers and area chair for their thoughtful feedback.
The authors declare no competing interests.
\end{ack}
\bibliographystyle{unsrt}
\bibliography{ref}

@article{lefevre2014survey,
  title={A survey on motion prediction and risk assessment for intelligent vehicles},
  author={Lef{\`e}vre, St{\'e}phanie and Vasquez, Dizan and Laugier, Christian},
  journal={ROBOMECH journal},
  volume={1},
  number={1},
  pages={1},
  year={2014},
  publisher={Springer}
}

@book{khalil2002nonlinear,
  title={Nonlinear systems},
  author={Khalil, Hassan K and Grizzle, Jessy W},
  volume={3},
  year={2002},
  publisher={Prentice Hall Upper Saddle River, NJ}
}

@inproceedings{allen2024hamiltonian,
  title={Hamiltonian GAN},
  author={Allen-Blanchette, Christine},
  booktitle={6th Annual Learning for Dynamics \& Control Conference},
  pages={1662--1674},
  year={2024},
  organization={PMLR}
}

@article{greydanus2019hamiltonian,
  title={Hamiltonian neural networks},
  author={Greydanus, Samuel and Dzamba, Misko and Yosinski, Jason},
  journal={Advances in neural information processing systems},
  volume={32},
  year={2019}
}

@article{mattheakis2022hamiltonian,
  title={Hamiltonian neural networks for solving equations of motion},
  author={Mattheakis, Marios and Sondak, David and Dogra, Akshunna S and Protopapas, Pavlos},
  journal={Physical Review E},
  volume={105},
  number={6},
  pages={065305},
  year={2022},
  publisher={APS}
}

@article{toth2019hamiltonian,
  title={Hamiltonian generative networks},
  author={Toth, Peter and Rezende, Danilo Jimenez and Jaegle, Andrew and Racani{\`e}re, S{\'e}bastien and Botev, Aleksandar and Higgins, Irina},
  journal={arXiv preprint arXiv:1909.13789},
  year={2019}
}

@article{cranmer2020lagrangian,
  title={Lagrangian neural networks},
  author={Cranmer, Miles and Greydanus, Sam and Hoyer, Stephan and Battaglia, Peter and Spergel, David and Ho, Shirley},
  journal={arXiv preprint arXiv:2003.04630},
  year={2020}
}

@article{allen2020lagnetvip,
  title={Lagnetvip: A lagrangian neural network for video prediction},
  author={Allen-Blanchette, Christine and Veer, Sushant and Majumdar, Anirudha and Leonard, Naomi Ehrich},
  journal={arXiv preprint arXiv:2010.12932},
  year={2020}
}

@article{sosanya2022dissipative,
  title={Dissipative hamiltonian neural networks: Learning dissipative and conservative dynamics separately},
  author={Sosanya, Andrew and Greydanus, Sam},
  journal={arXiv preprint arXiv:2201.10085},
  year={2022}
}

@article{zhong2020dissipative,
  title={Dissipative SymODEN: Encoding Hamiltonian dynamics with dissipation and control into deep learning},
  author={Zhong, Yaofeng Desmond and Dey, Biswadip and Chakraborty, Amit},
  journal={arXiv preprint arXiv:2002.08860},
  year={2020}
}

@article{zhong2019symplectic,
  title={Symplectic ode-net: Learning hamiltonian dynamics with control},
  author={Zhong, Yaofeng Desmond and Dey, Biswadip and Chakraborty, Amit},
  journal={arXiv preprint arXiv:1909.12077},
  year={2019}
}

@article{yoon2019time,
  title={Time-series generative adversarial networks},
  author={Yoon, Jinsung and Jarrett, Daniel and Van der Schaar, Mihaela},
  journal={Advances in neural information processing systems},
  volume={32},
  year={2019}
}

@article{chen2019symplectic,
  title={Symplectic recurrent neural networks},
  author={Chen, Zhengdao and Zhang, Jianyu and Arjovsky, Martin and Bottou, L{\'e}on},
  journal={arXiv preprint arXiv:1909.13334},
  year={2019}
}

@article{lishkova2023discrete,
  title={Discrete Lagrangian neural networks with automatic symmetry discovery},
  author={Lishkova, Yana and Scherer, Paul and Ridderbusch, Steffen and Jamnik, Mateja and Li{\`o}, Pietro and Ober-Bl{\"o}baum, Sina and Offen, Christian},
  journal={IFAC-PapersOnLine},
  volume={56},
  number={2},
  pages={3203--3210},
  year={2023},
  publisher={Elsevier}
}

@article{mason2023learning,
  title={Learning to predict 3D rotational dynamics from images of a rigid body with unknown mass distribution},
  author={Mason, Justice J and Allen-Blanchette, Christine and Zolman, Nicholas and Davison, Elizabeth and Leonard, Naomi Ehrich},
  journal={Aerospace},
  volume={10},
  number={11},
  pages={921},
  year={2023},
  publisher={MDPI}
}

@inproceedings{rodas2020re,
  title={Re-Hamiltonian Generative Networks},
  author={Rodas, Carles Balsells and Canal, Oleguer and Taschin, Federico},
  booktitle={ML Reproducibility Challenge 2020},
  year={2020}
}

@inproceedings{chen2021data,
  title={Data-driven prediction of general Hamiltonian dynamics via learning exactly-symplectic maps},
  author={Chen, Renyi and Tao, Molei},
  booktitle={International conference on machine learning},
  pages={1717--1727},
  year={2021},
  organization={PMLR}
}

@article{aboussalah2025geohnns,
  title={GeoHNNs: Geometric Hamiltonian Neural Networks},
  author={Aboussalah, Amine Mohamed and Ed-dib, Abdessalam},
  journal={arXiv preprint arXiv:2507.15678},
  year={2025}
}

@inproceedings{gordon2021latent,
  title={Latent neural differential equations for video generation},
  author={Gordon, Cade and Parde, Natalie},
  booktitle={NeurIPS 2020 Workshop on Pre-registration in Machine Learning},
  pages={73--86},
  year={2021},
  organization={PMLR}
}

@article{khan2022hamiltonian,
  title={Hamiltonian latent operators for content and motion disentanglement in image sequences},
  author={Khan, Asif and Storkey, Amos J},
  journal={Advances in Neural Information Processing Systems},
  volume={35},
  pages={7250--7263},
  year={2022}
}

@article{chen2018neural,
  title={Neural ordinary differential equations},
  author={Chen, Ricky TQ and Rubanova, Yulia and Bettencourt, Jesse and Duvenaud, David K},
  journal={Advances in neural information processing systems},
  volume={31},
  year={2018}
}

@book{marion2013classical,
  title={Classical dynamics of particles and systems},
  author={Marion, Jerry B},
  year={2013},
  publisher={Academic Press}
}

@article{matsubara2022finde,
  title={FINDE: Neural differential equations for finding and preserving invariant quantities},
  author={Matsubara, Takashi and Yaguchi, Takaharu},
  journal={arXiv preprint arXiv:2210.00272},
  year={2022}
}

@article{kasim2022constants,
  title={Constants of motion network},
  author={Kasim, Muhammad Firmansyah and Lim, Yi Heng},
  journal={Advances in Neural Information Processing Systems},
  volume={35},
  pages={25295--25305},
  year={2022}
}

@article{brunton2016discovering,
  title={Discovering governing equations from data by sparse identification of nonlinear dynamical systems},
  author={Brunton, Steven L and Proctor, Joshua L and Kutz, J Nathan},
  journal={Proceedings of the national academy of sciences},
  volume={113},
  number={15},
  pages={3932--3937},
  year={2016},
  publisher={National Academy of Sciences}
}

@misc{
unterthiner2019fvd,
title={{FVD}: A new Metric for Video Generation},
author={Thomas Unterthiner and Sjoerd van Steenkiste and Karol Kurach and Rapha{\"e}l Marinier and Marcin Michalski and Sylvain Gelly},
year={2019},
url={https://openreview.net/forum?id=rylgEULtdN}
}

@article{paredes2024output,
  title={Output-only identification of self-excited systems using discrete-time Lur'e models with application to a gas-turbine combustor},
  author={Paredes, Juan A and Yang, Yulong and Bernstein, Dennis S},
  journal={International Journal of Control},
  volume={97},
  number={2},
  pages={187--212},
  year={2024},
  publisher={Taylor \& Francis}
}

@inproceedings{richards2024output,
  title={Output-Only Identification of Lur'e Systems with Hysteretic Feedback Nonlinearities},
  author={Richards, Riley J and Yang, Yulong and Paredes, Juan A and Bernstein, Dennis S},
  booktitle={2024 American Control Conference (ACC)},
  pages={2891--2896},
  year={2024},
  organization={IEEE}
}

@article{yang2024learning,
  title={Learning Color Equivariant Representations},
  author={Yang, Yulong and O'Mahony, Felix and Allen-Blanchette, Christine},
  journal={arXiv preprint arXiv:2406.09588},
  year={2024}
}

@article{zhong2025gagrasp,
  title={Gagrasp: Geometric algebra diffusion for dexterous grasping},
  author={Zhong, Tao and Allen-Blanchette, Christine},
  journal={arXiv preprint arXiv:2503.04123},
  year={2025}
}

@article{kingma2014adam,
  title={Adam: A method for stochastic optimization},
  author={Kingma, Diederik P},
  journal={arXiv preprint arXiv:1412.6980},
  year={2014}
}

@article{yang2024behavior,
  title={Behavior-inspired neural networks for relational inference},
  author={Yang, Yulong and Feng, Bowen and Wang, Keqin and Leonard, Naomi Ehrich and Dieng, Adji Bousso and Allen-Blanchette, Christine},
  journal={arXiv preprint arXiv:2406.14746},
  year={2024}
}

@article{galioto2020bayesian2,
  title={Bayesian system ID: optimal management of parameter, model, and measurement uncertainty},
  author={Galioto, Nicholas and Gorodetsky, Alex Arkady},
  journal={Nonlinear Dynamics},
  volume={102},
  number={1},
  pages={241--267},
  year={2020},
  publisher={Springer}
}

@inproceedings{paredes2021identification,
  title={Identification of self-excited systems using discrete-time, time-delayed Lur'e models},
  author={Paredes, Juan and Bernstein, Dennis S},
  booktitle={2021 American Control Conference (ACC)},
  pages={3939--3944},
  year={2021},
  organization={IEEE}
}

@article{schmidt2009distilling,
  title={Distilling free-form natural laws from experimental data},
  author={Schmidt, Michael and Lipson, Hod},
  journal={science},
  volume={324},
  number={5923},
  pages={81--85},
  year={2009},
  publisher={American Association for the Advancement of Science}
}

@article{rubanova2019latent,
  title={Latent ordinary differential equations for irregularly-sampled time series},
  author={Rubanova, Yulia and Chen, Ricky TQ and Duvenaud, David K},
  journal={Advances in neural information processing systems},
  volume={32},
  year={2019}
}

@article{li2025latent,
  title={Latent Mixture of Symmetries for Sample-Efficient Dynamic Learning},
  author={Li, Haoran and Xiao, Chenhan and Guo, Muhao and Weng, Yang},
  journal={arXiv preprint arXiv:2510.03578},
  year={2025}
}

@article{abdi2010principal,
  title={Principal component analysis},
  author={Abdi, Herv{\'e} and Williams, Lynne J},
  journal={Wiley interdisciplinary reviews: computational statistics},
  volume={2},
  number={4},
  pages={433--459},
  year={2010},
  publisher={Wiley Online Library}
}

@article{o2024reconstructed,
  title={Reconstructed Performance of Mars 2020 Parachute Decelerator System},
  author={O’Farrell, Clara and Clark, Ian G},
  journal={Journal of Spacecraft and Rockets},
  volume={61},
  number={5},
  pages={1313--1328},
  year={2024},
  publisher={American Institute of Aeronautics and Astronautics}
}

@article{huh2020time,
  title={Time-reversal symmetric ode network},
  author={Huh, In and Yang, Eunho and Hwang, Sung Ju and Shin, Jinwoo},
  journal={Advances in Neural Information Processing Systems},
  volume={33},
  pages={19016--19027},
  year={2020}
}

@article{desai2021port,
  title={Port-Hamiltonian neural networks for learning explicit time-dependent dynamical systems},
  author={Desai, Shaan A and Mattheakis, Marios and Sondak, David and Protopapas, Pavlos and Roberts, Stephen J},
  journal={Physical Review E},
  volume={104},
  number={3},
  pages={034312},
  year={2021},
  publisher={APS}
}

@incollection{maschke1993port,
  title={Port-controlled Hamiltonian systems: modelling origins and systemtheoretic properties},
  author={Maschke, Bernhard M and van der Schaft, Arjan J},
  booktitle={Nonlinear control systems design 1992},
  pages={359--365},
  year={1993},
  publisher={Elsevier}
}

@book{james2021introduction,
  title={An introduction to statistical learning: with applications in R},
  author={James, Gareth and Witten, Daniela and Hastie, Trevor and Tibshirani, Robert and others},
  volume={2},
  year={2021},
  publisher={Springer}
}

@article{kaheman2023experimental,
  title={The experimental multi-arm pendulum on a cart: A benchmark system for chaos, learning, and control},
  author={Kaheman, Kadierdan and Fasel, Urban and Bramburger, Jason J and Strom, Benjamin and Kutz, J Nathan and Brunton, Steven L},
  journal={HardwareX},
  volume={15},
  pages={e00465},
  year={2023},
  publisher={Elsevier}
}

@article{jin2020sympnets,
  title={SympNets: Intrinsic structure-preserving symplectic networks for identifying Hamiltonian systems},
  author={Jin, Pengzhan and Zhang, Zhen and Zhu, Aiqing and Tang, Yifa and Karniadakis, George Em},
  journal={Neural Networks},
  volume={132},
  pages={166--179},
  year={2020},
  publisher={Elsevier}
}

@inproceedings{cho2014learning,
  title={Learning phrase representations using RNN encoder--decoder for statistical machine translation},
  author={Cho, Kyunghyun and Van Merri{\"e}nboer, Bart and Gul{\c{c}}ehre, {\c{C}}a{\u{g}}lar and Bahdanau, Dzmitry and Bougares, Fethi and Schwenk, Holger and Bengio, Yoshua},
  booktitle={Proceedings of the 2014 conference on empirical methods in natural language processing (EMNLP)},
  pages={1724--1734},
  year={2014}
}

@inproceedings{garcia2025learning,
  title={Learning physics from video: Unsupervised physical parameter estimation for continuous dynamical systems},
  author={Garcia, Alejandro Casta{\~n}eda and Warchocki, Jan and van Gemert, Jan and Brinks, Daan and Tomen, Nergis},
  booktitle={Proceedings of the Computer Vision and Pattern Recognition Conference},
  pages={27924--27933},
  year={2025}
}

@inbook{Jolliffe20114PCA,
author = {Jolliffe, Ian},
publisher = {John Wiley \& Sons, Ltd},
isbn = {9781118445112},
title = {Principal Component Analysis},
booktitle = {Wiley StatsRef: Statistics Reference Online},
chapter = {},
pages = {},
doi = {https://doi.org/10.1002/9781118445112.stat06472},
url = {https://onlinelibrary.wiley.com/doi/abs/10.1002/9781118445112.stat06472},
eprint = {https://onlinelibrary.wiley.com/doi/pdf/10.1002/9781118445112.stat06472},
year = {2014},
}

@inproceedings{cohen2016group,
  title={Group equivariant convolutional networks},
  author={Cohen, Taco and Welling, Max},
  booktitle={International conference on machine learning},
  pages={2990--2999},
  year={2016},
  organization={PMLR}
}

@article{towers2024gymnasium,
  title={Gymnasium: A standard interface for reinforcement learning environments},
  author={Towers, Mark and Kwiatkowski, Ariel and Terry, Jordan and Balis, John U and De Cola, Gianluca and Deleu, Tristan and Goul{\~a}o, Manuel and Kallinteris, Andreas and Krimmel, Markus and KG, Arjun and others},
  journal={arXiv preprint arXiv:2407.17032},
  year={2024}
}

@inproceedings{yuan2024diffusion,
  title={Diffusion-ts: Interpretable diffusion for general time series generation},
  author={Yuan, Xinyu and Qiao, Yan},
  booktitle={International Conference on Learning Representations},
  volume={2024},
  pages={41582--41610},
  year={2024}
}

@inproceedings{naiman2024generative,
  title={Generative modeling of regular and irregular time series data via Koopman VAEs},
  author={Naiman, Ilan and Erichson, N Benjamin and Ren, Pu and Mahoney, Michael W and Azencot, Omri},
  booktitle={International conference on learning representations},
  volume={2024},
  pages={50547--50573},
  year={2024}
}

@inproceedings{brenner2022tractable,
  title={Tractable dendritic RNNs for reconstructing nonlinear dynamical systems},
  author={Brenner, Manuel and Hess, Florian and Mikhaeil, Jonas M and Bereska, Leonard F and Monfared, Zahra and Kuo, Po-Chen and Durstewitz, Daniel},
  booktitle={International conference on machine learning},
  pages={2292--2320},
  year={2022},
  organization={Pmlr}
}

@article{wu2026f3no,
  title={F$^{3}$NO: Frequency-Decomposed Finite-Time Flow-map Neural Operators with Cross-Scale Conditioning},
  author={Wu, Fan and Jing, Cheng and Lee, Kookjin},
  journal={arXiv preprint arXiv:2610.10998},
  year={2026}
}

@inproceedings{wu2024identifying,
  title={Identifying Contemporaneous and Lagged Dependence Structures by Promoting Sparsity in Continuous-time Neural Networks},
  author={Wu, Fan and Cho, Woojin and Korotky, David and Hong, Sanghyun and Rim, Donsub and Park, Noseong and Lee, Kookjin},
  booktitle={Proceedings of the 33rd ACM International Conference on Information and Knowledge Management},
  pages={2534--2543},
  year={2024}
}
\newpage
\appendix
%
\section{Dataset and Training}\label{apn:dataset}
This appendix documents the datasets and training details used in Section~\ref{sec:results}, covering both simulated and real-world systems.
The simulated data comes from two sources: five conservative systems (mass-spring oscillator, ideal pendulum, double pendulum, two-body, and three-body) and two non-conservative systems (damped pendulum and forced Duffing oscillator) generated using the HGN re-implementation pipeline~\citep{rodas2020re}.
%
%
\subsection{Dataset Generation}
All simulated trajectories and video are produced via integration with timestep $\Delta t=0.05$ and length $T=30$ frames.
For the long-horizon rollout example in Table~\ref{tab:result_distributional_fidelity}, the double pendulum is simulated to $T=200$ frames.
The real-world data comprises pendulum trajectory recordings~\citep{schmidt2009distilling}, double pendulum trajectory recordings~\citep{kaheman2023experimental}, and double pendulum video recordings~\citep{kaheman2023experimental}.
We detail the equations of motion, initial conditions, and sampling procedures for each system presented in Section~\ref{sec:results}.
%
%
\subsubsection{Mass-Spring System.}
The mass-spring oscillator is governed by the Hamiltonian
\begin{equation}
    \mathcal{H} = \frac{p^{2}}{2m} + \frac{1}{2}ky^{2},
\end{equation}
with dynamics
\begin{equation}
    \ddot y = -2c\sqrt{k/m}\dot y - (k/m)y.
\end{equation}
Our experiments use mass $m=0.5$ and elastic constant $k=2.0$.
The initial position $y_{0}$ is sampled uniformly from $\left[0.1, 1.0\right]$.
As the mass-spring is a one-dimensional system, we pad with $x=0.0$ to embed the trajectory in the 2D Cartesian frame used by the other systems.
%
%
\subsubsection{Duffing Oscillator.}
The conservative Duffing oscillator is governed by the Hamiltonian
\begin{equation}
    \mathcal{H} = \frac{1}{2}p^{2} + \frac{1}{2}\alpha q^{2} + \frac{1}{4}\beta q^{4},
\end{equation}
with dynamics of the forced case being
\begin{equation}
    \ddot{x} = \gamma\cos\omega t - \delta\dot{x} - \alpha x - \beta x^{3}.
\end{equation}
Our experiments use forcing $\delta=0.15$, $\alpha=-0.2$, and $\beta=0.2$.
The initial position $x_{0}$ is sampled uniformly from $\left[0.2, 1.0\right]$.
As the Duffing oscillator is a one-dimensional system, we pad with $y=0.0$ to embed the trajectory in the 2D Cartesian frame used by the other systems.
%
%
\subsubsection{Ideal Pendulum.}
The ideal pendulum system is governed by the Hamiltonian 
\begin{equation}
    \mathcal{H} = \frac{p^2}{2mL^2} + mgL(1-\cos\theta),
\end{equation}
with dynamics
\begin{equation}
    \ddot\theta = -(g/L)\sin\theta.
\end{equation}
Our experiments use mass $m=0.5$, length $L=1.0$, and gravity $g=9.81$. 
The initial angular position is sampled uniformly from $\left[-\pi, \pi\right]$.
While the system is simulated using canonical coordinates, the final training trajectory is projected to Cartesian coordinates or video sequences.
%
%
\subsubsection{Double Pendulum.}
The chaotic double pendulum system is represented by the Hamiltonian
\begin{equation}
    \mathcal{H} = \frac{1}{2mL^{2}} \cdot \frac{p_{1}^{2} + 2p_{2}^{2} - 2p_{1}p_{2}\cos(\theta_{1}-\theta_{2})}{1+\sin^{2}(\theta_{1}-\theta_{2})} + mgL(3-2\cos\theta_{1}-\cos\theta_{2}),
\end{equation}
where $\theta_{1},\theta_{2}$ denote the angular positions and $p_{1},p_{2}$ denote the corresponding conjugate momenta. The dynamics are given by Hamilton's equations. Let $\Delta = \theta_{1}-\theta_{2}$, $D = 1+\sin^{2}(\Delta)$, and $N = p_{1}^{2}+2p_{2}^{2}-2p_{1}p_{2}\cos(\Delta)$. Then,
\begin{align}
    \dot{\theta}_{1} &= \frac{p_{1}-p_{2}\cos(\Delta)}{mL^{2}D},\\
    \dot{\theta}_{2} &= \frac{2p_{2}-p_{1}\cos(\Delta)}{mL^{2}D}, \\
    \dot{p}_{1} &= \frac{\sin(\Delta)(N\cos(\Delta)-p_{1}p_{2}D)}{mL^{2}D^{2}} - 2mgL\sin\theta_{1},\\
    \dot{p}_{2} &= \frac{\sin(\Delta)(p_{1}p_{2}D-N\cos(\Delta))}{mL^{2}D^{2}} - mgL\sin\theta_{2},
\end{align}
with parameters $m=1.0$, $L=1.0$, and $g=3.0$.
The initial angular position is sampled uniformly from $\left[-0.5\pi, 0.5\pi\right]$.
While the system is simulated using canonical coordinates, the final training trajectory is projected to Cartesian coordinates or video sequences.
%
%
\subsubsection{Planar Two-Body System.}
The gravitational two-body system is represented by the Hamiltonian
\begin{equation}
    \mathcal{H} = \frac{|\mathbf{p}_1|^2}{2m_1} + \frac{|\mathbf{p}_2|^2}{2m_2} - \frac{Gm_1m_2}{|\mathbf{q}_1-\mathbf{q}_2|},
\end{equation}
where the dynamics are governed by 
\begin{equation}
    m_i\ddot{\mathbf{q}}_i = G\sum_{j \neq i} \frac{m_im_j(\mathbf{q}_j - \mathbf{q}_i)}{|\mathbf{q}_j - \mathbf{q}_i|^3},
\end{equation}
with equal masses $m=1.0$ and gravitational constant $G=1.0$. 
The two objects are placed on the same circular orbit at opposite position with the radius ranging from 0.5 to 1.5.
Their initial velocities are set perpendicular to the radius vector, and a small perturbations of magnitude $0.1$ is added to the velocities.
%
%
\subsubsection{Planar Three-Body System.}
The gravitational three-body system is represented by the Hamiltonian
\begin{equation}
    \mathcal{H} = \sum_{i=1}^{3}\frac{|\mathbf{p}_i|^2}{2m_i} - \sum_{i<j}\frac{Gm_im_j}{|\mathbf{q}_i-\mathbf{q}_j|},
\end{equation}
with equal masses $m=1.0$ and gravitational constant $G=1.0$. 
Three equal mass objects are initialized in an equilateral triangle configuration with $120^{\circ}$ angular separation within radius bounds $[0.9, 1.5]$. 
Their initial velocities are set perpendicular to the radius vector, and a small perturbations of magnitude $0.1$ is added to the velocities.
%
%
\subsection{Training Setting}
All training of SPS-GAN was performed on cluster nodes each with 1 Nvidia L40 GPU, 16 core partition of an Intel Xeon Gold 5320, and 64G of DDR4 3200MHz RDIMM.

\textbf{SPS-GAN trajectory training hyperparameter.}
SPS-GAN for generating time-series data is trained through 20000 steps with batch size of 128 (pendulum, double pendulum, mass-spring, two-body) or 160 (three-body).
It is optimized using Adam~\citep{kingma2014adam} with $\beta_{1}=0.3$ and $\beta_{2}=0.999$ using a learning rate of $5\times10^{-5}$.
The phase-space map and HNN module is initialized as an MLP with hidden size $100$ using ReLU activation.
The trajectory generator is initialized as an MLP with hidden size $512$ using SoftPlus activation.
We set the latent motion space to $d_{\mathrm{m}}=10$, content space to $d_{\mathrm{c}}=0$, and output size $d_{\mathrm{output}}=10$.
For training we use trajectories of length $30$; for discrimination we set $\ell=16$.

\textbf{SPS-GAN video training hyperparameter.}
SPS-GAN for generating video data is trained through 20000 steps with batch size of 16.
It is optimized using Adam with $\beta_{1}=0.3$ and $\beta_{2}=0.999$ using a learning rate of $5\times10^{-5}$.
The phase-space map and pHNN module is initialized as an MLP with hidden size $100$ using ReLu activation.
The image generator is initialized as an CNN with 32 filter and 3 channel.
We set the latent motion space to $d_{\mathrm{m}}=10$, content space to $d_{\mathrm{c}}=50$.
For training we use video sequences of length $30$; for discrimination we set $\ell=16$.
%
%
\section{Results}\label{apn:results}
\begin{figure}[t]
    \begin{subfigure}[t]{\textwidth}
        \centering
        \includegraphics[width=0.9\columnwidth]{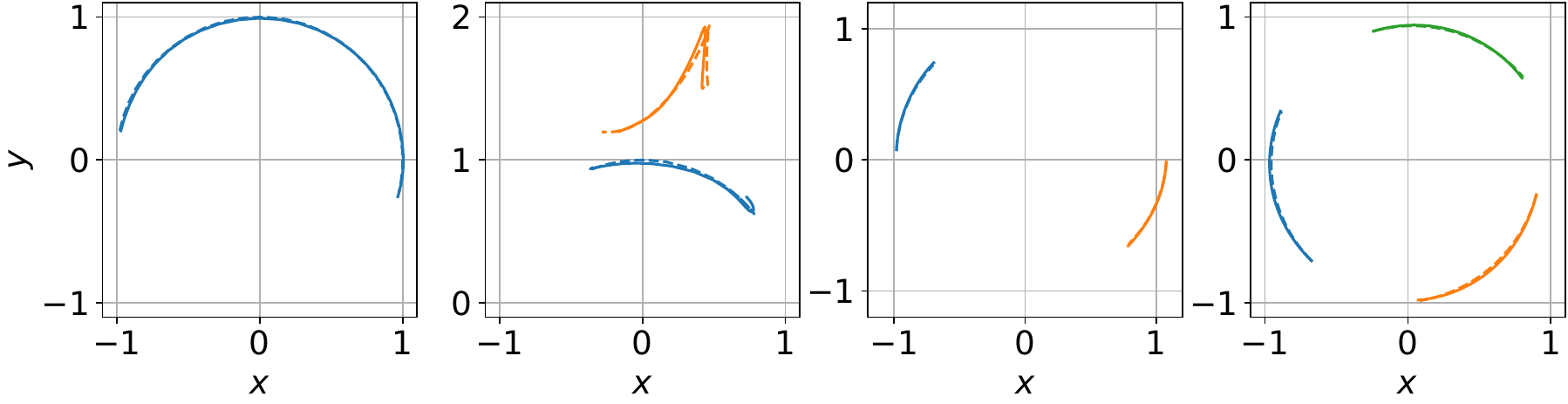}
        \vskip -0.06in
        \caption{\textbf{SPS-GAN-1 unsupervised modeling of single systems.}}
        \label{fig:SPSGAN_ConstantSystemTrajComp}
    \end{subfigure}
    \vskip 0.04in
    \begin{subfigure}[t]{\textwidth}
        \centering
        \includegraphics[width=0.9\columnwidth]{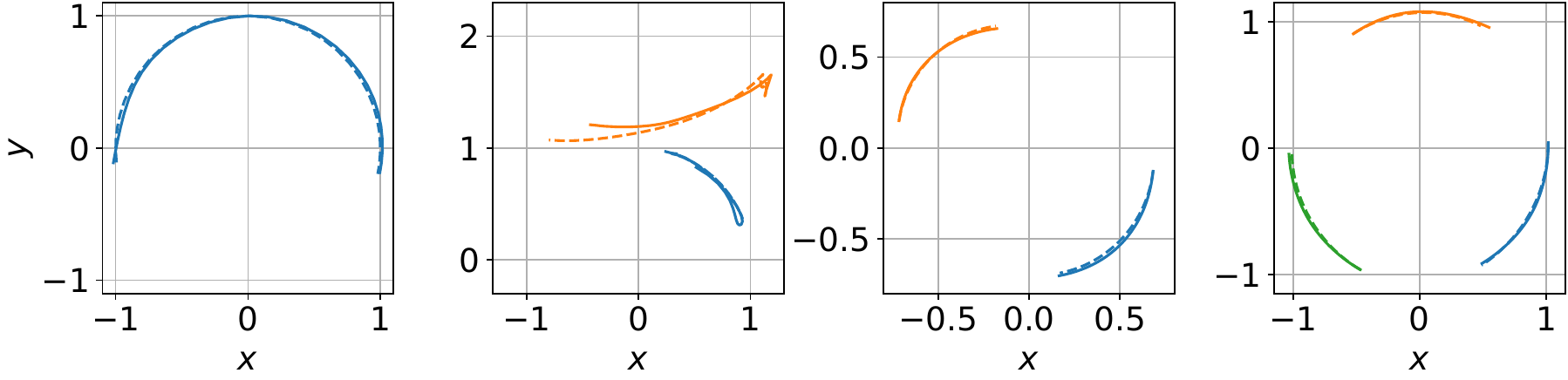}
        \vskip -0.06in
        \caption{\textbf{SPS-GAN-5 joint modeling of five systems.}}
        \label{fig:SPSGAN_VariableSystemTrajComp}
    \end{subfigure}
    \caption{\textbf{Example Cartesian trajectory rollout.} Comparison between ground-truth (dotted) and generated (solid) trajectories across the single pendulum, double pendulum, two-body, and three-body systems. Whether generating dynamics from \textbf{(a)} one system at a time or \textbf{(b)} multiple systems jointly, SPS-GAN trajectories closely follow the reference dynamics.}
    \label{fig:CHGAN_LowDimComp}
    \vspace{-0.1in}
\end{figure}
\begin{figure}[b]
    \vspace{-0.1in}
    \centering
    \includegraphics[width=0.75\textwidth]{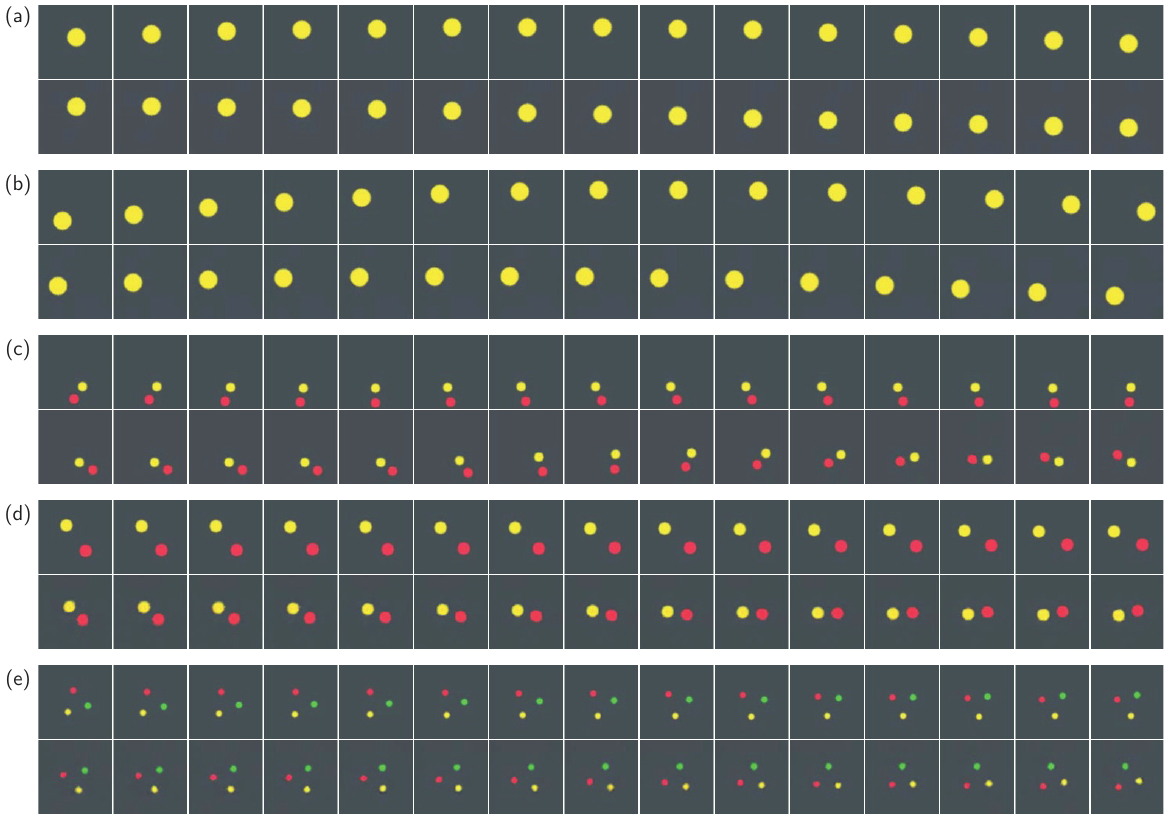}
    \vskip -0.06in
    \caption{\textbf{Example video rollout under fixed parameters.} Real system (top) and generated video (bottom) for (a) mass-spring oscillator, (b) ideal pendulum, (c) double pendulum, (d) two-body, and (e) three-body system, all under fixed parameters.}
    \label{fig:CHGAN_TMLR_CCCComparison}
\end{figure}
We expand on the analysis in Section~\ref{sec:results} with example trajectory and video rollouts, extended ablation analysis, robustness of SPS-GAN to the choice of latent dimension $d_{\mathrm{lat}}$, and analytical derivations of the degrees of freedom and canonical coordinates for each system.
%
%
\subsection{Example Cartesian Trajectory Rollout}\label{subsec:result_traj_rollout}
We include example Cartesian trajectory rollouts generated by SPS-GAN in Figure~\ref{fig:CHGAN_LowDimComp} for two training regimes: SPS-GAN-1, trained without supervision on one system at a time, and SPS-GAN-5, trained jointly on all five systems with supervisory conditional labels.

For systems reported in Table~\ref{tab:result_nonconservative_accuracy}, the learnable dissipation $\mathbf{D}_{\phi}$ parameter and input map $\mathbf{F}_{\psi}$ accommodates the non-conservative dynamics in Equation~\eqref{eqn:port_ham}.
We constraint the dissipation term $\mathbf{D}_{\phi}$ to be positive semidefinite through Cholesky decomposition such that $\mathbf{D}_{\phi} = \tilde{\mathbf{D}}_{\phi}\tilde{\mathbf{D}}_{\phi}^{\top}$.
%
%
\subsection{Example Video Rollout}\label{subsec:result_video_rollout}
\begin{figure}[t]
    \centering
    \includegraphics[width=0.75\columnwidth]{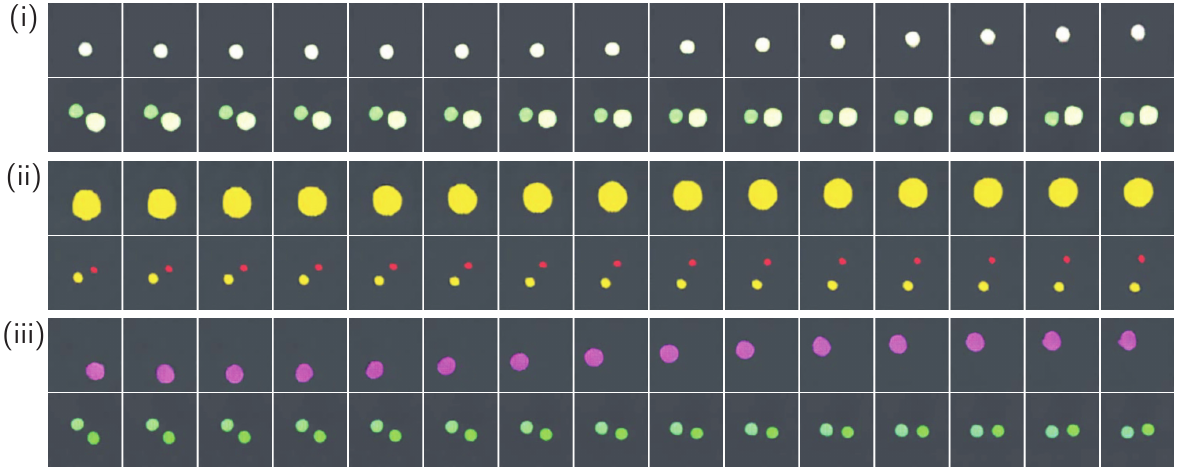}
    \vskip -0.06in
    \caption{\textbf{Generated video under color and physics variation.} SPS-GAN-5 generated video of mass-spring and two body systems with \textbf{(i)} physics and color varied (FVD 182.15), \textbf{(ii)} only physics varied (FVD 135.38), and \textbf{(iii)} only color varied (FVD 135.25). The default color setting uses yellow, red, and green.}
    \label{fig:CHGAN_TMLR_VVVComparison}
    \vspace{-0.15in}
\end{figure}

We include example video rollouts generated by SPS-GAN in Figure~\ref{fig:CHGAN_TMLR_CCCComparison}, trained without supervision on one system at a time with no color or physics variation. 
We include example video rollouts generated by SPS-GAN-5 jointly modeling all five systems in Figure~\ref{fig:CHGAN_TMLR_VVVComparison}, trained with conditional labels encoding variations in color and physics.
%
%
\subsection{Controlled Pendulum}\label{subsec:result_control_pend}
\begin{figure}[b]
    \vspace{-0.15in}
    \centering
    \includegraphics[width=0.9\columnwidth]{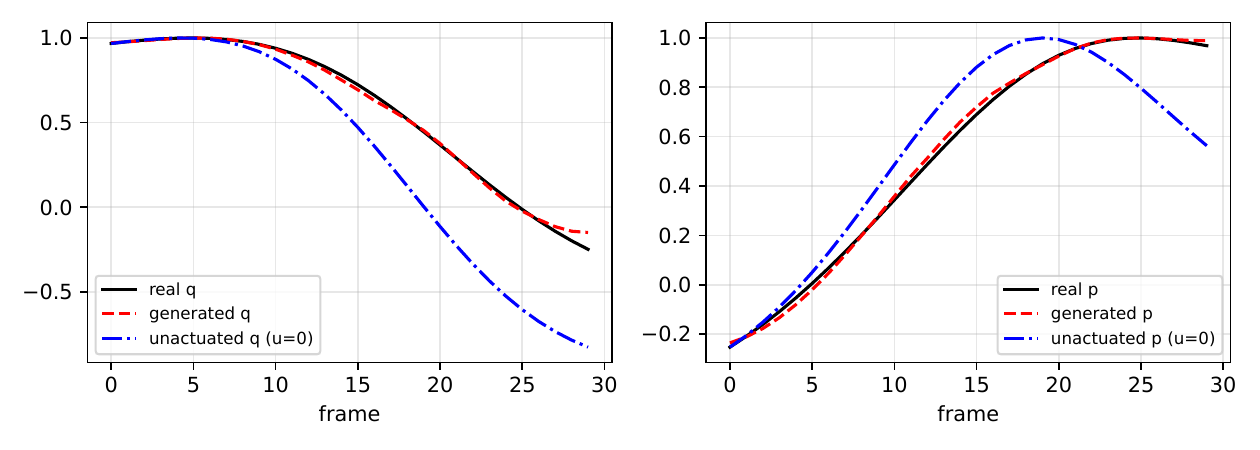}
    \vskip -0.14in
    \caption{\textbf{Accuracy of generated trajectory for actuated pendulum.} SPS-GAN generated trajectory under constant external actuation closely matches the reference dynamics from gymnasium\citep{towers2024gymnasium}.}
    \label{fig:SPSGAN_ControlledPendulum}
\end{figure}
We include additional results on the generated trajectory of unsupervised SPS-GAN on a controlled pendulum, using the Gymnasium environment~\citep{towers2024gymnasium}.
At test time SPS-GAN generates a trajectory under a constant input.
The first coordinate of the generated trajectory and input are used to create the ground truth reference using the gymnasium environment.
We show that SPS-GAN is able to accurate learned the dynamics under external actuation in Figure~\ref{fig:SPSGAN_ControlledPendulum}, where the generated trajectory closely matches the gymnasium simulated one.
Notably the SPS-GAN generated trajectory deviates significantly from the unactuated dynamics, indicating that SPS-GAN is able to learn the dynamics of actuation rather than small perturbations.
Empirically, SPS-GAN generated controlled pendulum yielded MSE of $4.91\times10^{-4}$, on par with the accuracy of unactuated ideal pendulum (see Table~\ref{tab:result_conservative_accuracy}).
%
%
\subsection{Analytical Derivation}\label{subsec:DoF_canonical_coordinates}
We show the analytical derivation of the degrees of freedom and canonical coordinates for the systems studied in Section~\ref{sec:results}, and verify that the identified degrees of freedom and recovered canonical coordinates match the analytical results.

\textbf{Mass Spring.} The mass-spring oscillator is described by a single generalized coordinate $y$ (the displacement from equilibrium), giving one degree of freedom by definition.

\textbf{Duffing Oscillator.} The Duffing oscillator is described by a single generalized coordinate $x$ (the displacement), giving one degree of freedom by definition.

\textbf{Single Pendulum.} The Cartesian coordinate $(x,y)$ can be parameterized by the angular position $\theta$
\begin{equation}
    x = L\sin\theta, \quad\quad y = -L\cos\theta,
\end{equation}
where $L$ is the length of the rod. 
The kinetic and potential energies, along with damping are
\begin{align}
    T &= \frac{1}{2}mL^{2}\dot{\theta}^{2},\\
    V &= -mgL\cos\theta,\\
    \mathcal{D} &= \frac{1}{2}c\dot{\theta}^{2},
\end{align}
where $c$ is the damping coefficient.
Therefore the unactuated Euler-Lagrange equation of Lagrangian $\mathcal{L}=T-V$ with damping becomes
\begin{equation}
    \frac{d}{dt}\left(\frac{\partial\mathcal{L}}{\partial\dot{\theta}}\right) - \frac{\partial\mathcal{L}}{\partial\theta} + \frac{\partial\mathcal{D}}{\partial\dot{\theta}} = 0.
\end{equation}
Solving and expanding the terms yields the dynamics equation for the single pendulum
\begin{equation}
    \ddot{\theta} = -\frac{c}{mL^{2}}\dot{\theta} - \frac{g}{L}\sin\theta,
\end{equation}
which only dependents on canonical coordinate $\theta$, showing a one degree of freedom system.

\textbf{Double Pendulum.} The Cartesian coordinates of the two masses can be parameterized by the angular positions $\theta_{1}$ and $\theta_{2}$ as
\begin{alignat}{3}
    x_{1} &= L_{1}\sin\theta_{1}, &&y_{1} = -L_{1}\cos\theta_{1},\\
    x_{2} &= L_{1}\sin\theta_{1} + L_{2}\sin\theta_{2}, \quad\quad &&y_{2} = -L_{1}\cos\theta_{1} - L_{2}\cos\theta_{2},
\end{alignat}
where $L_{1}$ and $L_{2}$ are the lengths of the two rods. The angles are measured from the downward vertical, with $\theta_{1}$ describing the first rod and $\theta_{2}$ describing the second rod. Let
\begin{equation}
    \Delta = \theta_{1} - \theta_{2}.
\end{equation}
Using the coordinate parameterization above, the squared speeds of the two masses are
\begin{align}
    v_{1}^{2} &= L_{1}^{2}\dot{\theta}_{1}^{2},\\
    v_{2}^{2} &= L_{1}^{2}\dot{\theta}_{1}^{2} + L_{2}^{2}\dot{\theta}_{2}^{2} + 2L_{1}L_{2}\cos\Delta\,\dot{\theta}_{1}\dot{\theta}_{2}.
\end{align}
The kinetic and potential energies, along with damping, are therefore
\begin{align}
    T &= \frac{1}{2}(m_{1}+m_{2})L_{1}^{2}\dot{\theta}_{1}^{2} + \frac{1}{2}m_{2}L_{2}^{2}\dot{\theta}_{2}^{2} + m_{2}L_{1}L_{2}\cos\Delta\,\dot{\theta}_{1}\dot{\theta}_{2},\\
    V &= -(m_{1}+m_{2})gL_{1}\cos\theta_{1} - m_{2}gL_{2}\cos\theta_{2},\\
    \mathcal{D} &= \frac{1}{2}c_{1}\dot{\theta}_{1}^{2} + \frac{1}{2}c_{2}(\dot{\theta}_{2}-\dot{\theta}_{1})^{2},
\end{align}
where $c_{1}$ is the damping coefficient at the first pivot, and $c_{2}$ is the damping coefficient at the second joint. The second damping term depends on the relative angular velocity between the two rods.
The Lagrangian is $\mathcal{L}=T-V$, and the unactuated Euler-Lagrange equations with damping are
\begin{equation}
    \frac{d}{dt}\left(\frac{\partial\mathcal{L}}{\partial\dot{\theta}_{i}}\right) - \frac{\partial\mathcal{L}}{\partial\theta_{i}} + \frac{\partial\mathcal{D}}{\partial\dot{\theta}_{i}} = 0, \quad i=1,2.
\end{equation}
For compactness, define
\begin{align}
    A &= (m_{1}+m_{2})L_{1}^{2}, \quad\quad B = m_{2}L_{1}L_{2}, \quad\quad C = m_{2}L_{2}^{2},\\
    G_{1} &= (m_{1}+m_{2})gL_{1}, \quad\quad G_{2} = m_{2}gL_{2}.
\end{align}
Solving and expanding the terms yields the coupled dynamics of the damped double pendulum,
\begin{align}
    A\ddot{\theta}_{1} + B\cos\Delta\,\ddot{\theta}_{2} + B\sin\Delta\,\dot{\theta}_{2}^{2} + G_{1}\sin\theta_{1} + (c_{1}+c_{2})\dot{\theta}_{1} - c_{2}\dot{\theta}_{2} &= 0,\\
    B\cos\Delta\,\ddot{\theta}_{1} + C\ddot{\theta}_{2} - B\sin\Delta\,\dot{\theta}_{1}^{2} + G_{2}\sin\theta_{2} - c_{2}\dot{\theta}_{1} + c_{2}\dot{\theta}_{2} &= 0.
\end{align}
Equivalently, the system can be written explicitly in terms of the angular accelerations. Define
\begin{align}
    R_{1} &= -B\sin\Delta\,\dot{\theta}_{2}^{2} - G_{1}\sin\theta_{1} - (c_{1}+c_{2})\dot{\theta}_{1} + c_{2}\dot{\theta}_{2},\\
    R_{2} &= B\sin\Delta\,\dot{\theta}_{1}^{2} - G_{2}\sin\theta_{2} + c_{2}\dot{\theta}_{1} - c_{2}\dot{\theta}_{2}.
\end{align}
Then the equations of motion are
\begin{align}
    \ddot{\theta}_{1} &= \frac{CR_{1} - B\cos\Delta\,R_{2}}{AC - B^{2}\cos^{2}\Delta},\\
    \ddot{\theta}_{2} &= \frac{-B\cos\Delta\,R_{1} + AR_{2}}{AC - B^{2}\cos^{2}\Delta}.
\end{align}
Thus, the damped double pendulum depends on the two canonical coordinates $\theta_{1}$ and $\theta_{2}$, showing a two-degree-of-freedom system.

\textbf{Planar Two-Body System.} Consider two point masses $m_{1}$ and $m_{2}$ moving in a plane under their mutual gravitational attraction. Let $(x_{1},y_{1})$ and $(x_{2},y_{2})$ denote the Cartesian positions of the two bodies, and define the relative coordinate from body $1$ to body $2$ as
\begin{equation}
    x = x_{2}-x_{1}, \quad\quad y = y_{2}-y_{1}.
\end{equation}
The relative coordinate can be parameterized in polar form by the radius $r$ between the bodies and the angle $\theta$ as
\begin{equation}
    x = r\cos\theta, \quad\quad y = r\sin\theta, \quad\quad r = \sqrt{x^{2}+y^{2}}.
\end{equation}
Let $M=m_{1}+m_{2}$ be the total mass and let $\mu=\frac{m_{1}m_{2}}{m_{1}+m_{2}}$ be the reduced mass. After separating out the center-of-mass motion, the kinetic and potential energies of the relative system are
\begin{align}
    T &= \frac{1}{2}\mu\left(\dot{r}^{2}+r^{2}\dot{\theta}^{2}\right),\\
    V &= -\frac{Gm_{1}m_{2}}{r},
\end{align}
where $G$ is the gravitational constant. Therefore the two-body problem can be written as an equivalent one-body problem with reduced mass $\mu$ moving in a central potential. The Lagrangian is
\begin{equation}
    \mathcal{L}=T-V=\frac{1}{2}\mu\left(\dot{r}^{2}+r^{2}\dot{\theta}^{2}\right)+\frac{Gm_{1}m_{2}}{r}.
\end{equation}
The unactuated Euler-Lagrange equations are
\begin{equation}
    \frac{d}{dt}\left(\frac{\partial\mathcal{L}}{\partial\dot{q}_{i}}\right)-\frac{\partial\mathcal{L}}{\partial q_{i}}=0, \quad\quad q_{i}\in\{r,\theta\}.
\end{equation}
Since the Lagrangian does not explicitly depend on $\theta$, the angular coordinate is cyclic, so its conjugate momentum is conserved
\begin{equation}
    \frac{\partial\mathcal{L}}{\partial\dot{\theta}}=\mu r^{2}\dot{\theta}=\ell,
\end{equation}
where $\ell$ is the conserved angular momentum of the reduced system. Equivalently, defining the specific angular momentum $h=\frac{\ell}{\mu}$ gives
\begin{equation}
    r^{2}\dot{\theta}=h, \quad\quad \dot{\theta}=\frac{h}{r^{2}}.
\end{equation}
The Euler-Lagrange equation for the radial coordinate is
\begin{equation}
    \frac{d}{dt}\left(\frac{\partial\mathcal{L}}{\partial\dot{r}}\right)-\frac{\partial\mathcal{L}}{\partial r}=0.
\end{equation}
Expanding the terms gives
\begin{equation}
    \mu\ddot{r}-\mu r\dot{\theta}^{2}+\frac{Gm_{1}m_{2}}{r^{2}}=0.
\end{equation}
Solving for $\ddot{r}$ yields
\begin{equation}
    \ddot{r}=r\dot{\theta}^{2}-\frac{G(m_{1}+m_{2})}{r^{2}}.
\end{equation}
Using conservation of angular momentum to eliminate $\dot{\theta}$, the radial equation of motion becomes
\begin{equation}
    \ddot{r}=\frac{h^{2}}{r^{3}}-\frac{G(m_{1}+m_{2})}{r^{2}}.
\end{equation}
Thus, after removing the center-of-mass motion, the relative two-body system has two degrees of freedom $\left(r,\theta\right)$.
Conservation of angular momentum renders $\theta$ cyclic and recoverable by quadrature from $\dot\theta = h/r^{2}$, reducing the dynamics to an effective one-degree-of-freedom system in the radial coordinate $r$.

\textbf{Planar Three-Body Problem.} We denote by $q_{j}\in\mathbb{R}^{2}$ the position vector of the $j$th particle of mass $m_{j}$, where $j=1,2,3$. The collision set is defined as
\begin{equation}
    \Delta_{ij} = \{q=(q_{1},q_{2},q_{3})\in(\mathbb{R}^{2})^{3}:q_{i}=q_{j}\}, \quad\quad \Delta = \cup_{1\leq i<j\leq 3}\Delta_{ij}.
\end{equation}
We consider $q=(q_{1},q_{2},q_{3})\in(\mathbb{R}^{2})^{3}\setminus\Delta$ in the center-of-mass frame, so that $\sum_{j=1}^{3}m_{j}q_{j}=0$. Let $M=\operatorname{diag}(m_{1},m_{1},m_{2},m_{2},m_{3},m_{3})$. With the gravitational constant absorbed into the units, the Newtonian potential and equations of motion are
\begin{equation}
    U(q) = \sum_{1\leq i<j\leq 3}\frac{m_{i}m_{j}}{\lVert q_{i}-q_{j}\rVert}, \quad\quad M\ddot q = \nabla U(q).
\end{equation}
We restrict to equilateral homographic solutions. Identifying $\mathbb{R}^{2}$ with $\mathbb{C}$, write
\begin{equation}
    q_{j}(t) = z(t)q_{j}^{(c)}, \quad\quad j=1,2,3,
\end{equation}
where $q^{(c)}=(q_{1}^{(c)},q_{2}^{(c)},q_{3}^{(c)})$ is a fixed equilateral configuration satisfying $\sum_{j=1}^{3}m_{j}q_{j}^{(c)}=0$, and $z(t)\in\mathbb{C}$ controls the global scale and rotation of the triangle. The corresponding homographic initial velocities satisfy
\begin{equation}
    \dot q_{j}(0) = \dot z(0)q_{j}^{(c)}, \quad\quad j=1,2,3.
\end{equation}
For the equal-mass case, this corresponds to initializing the three bodies at the vertices of an equilateral triangle and giving them equal velocity magnitudes separated by $120^{\circ}$. For example, if $q_{j}^{(c)}=\rho e^{\mathrm{i}(\alpha_{0}+2\pi(j-1)/3)}$ and $\dot z(0)=\frac{v}{\rho}e^{\mathrm{i}\beta_{0}}$, then
\begin{equation}
    \dot q_{j}(0) = v e^{\mathrm{i}(\alpha_{0}+\beta_{0}+2\pi(j-1)/3)}, \quad\quad j=1,2,3,
\end{equation}
so the velocity vectors have equal magnitude and are separated by $120^{\circ}$.
Let $a_{c}=\lVert q_{i}^{(c)}-q_{j}^{(c)}\rVert$ denote the side length of the reference equilateral triangle. Since $q^{(c)}$ is equilateral and centered at the center of mass,
\begin{equation}
    \nabla_{q_{j}}U(q^{(c)}) = \sum_{k\neq j}\frac{m_{j}m_{k}(q_{k}^{(c)}-q_{j}^{(c)})}{a_{c}^{3}} = -\frac{m_{j}m_{\mathrm{tot}}}{a_{c}^{3}}q_{j}^{(c)}, \quad\quad m_{\mathrm{tot}}=m_{1}+m_{2}+m_{3}.
\end{equation}
Thus $q^{(c)}$ is a central configuration satisfying
\begin{equation}
    \nabla U(q^{(c)}) = -\lambda Mq^{(c)}, \quad\quad \lambda = \frac{m_{\mathrm{tot}}}{a_{c}^{3}}.
\end{equation}
Using the homogeneity of the Newtonian potential, the gradient satisfies
\begin{equation}
    \nabla U(zq^{(c)}) = \frac{z}{\lvert z\rvert^{3}}\nabla U(q^{(c)}).
\end{equation}
This central-configuration condition is the structural reason the homographic (i.e., the shape is preserved, only rotation and scaling is applied) approach is consistent, as substituting $q(t)=z(t)q^{(c)}$ into Newton's equations produces a right-hand side proportional to $\nabla U(q^{(c)})$, and only when this gradient is parallel to $Mq^{(c)}$ does the system collapse to a scalar equation in $z(t)$.
For a non-central reference configuration, the gradient would have components transverse to $Mq^{(c)}$, the homographic form would fail to be preserved in time, and the reduction would break down.
Substituting $q(t)=z(t)q^{(c)}$ into Newton's equations gives
\begin{equation}
    \ddot z(t)Mq^{(c)} = \frac{z(t)}{\lvert z(t)\rvert^{3}}\nabla U(q^{(c)}) = -\lambda\frac{z(t)}{\lvert z(t)\rvert^{3}}Mq^{(c)}.
\end{equation}
Therefore, the three-body dynamics restricted to the equilateral homographic subspace reduce to the scalar complex equation
\begin{equation}
    \ddot z = -\lambda\frac{z}{\lvert z\rvert^{3}}.
\end{equation}
Let
\begin{equation}
    z(t)=r(t)e^{\mathrm{i}\phi(t)}.
\end{equation}
Here $r(t)$ controls the size of the equilateral triangle, while $\phi(t)$ controls its global rotation. In particular, the physical side length is
\begin{equation}
    a(t)=a_{c}r(t).
\end{equation}
Thus, even if the reference configuration or units are normalized so that $\lambda=1$, the equilateral size is still allowed to vary through the dynamic variable $r(t)$. Expanding $\ddot z$ gives
\begin{table}[b]
    \vspace{-0.15in}
    \caption{\textbf{Identifying system degrees of freedom.} SPS-GAN correctly recovers the degrees of freedom for conservative and non-conservative systems, under observations of Cartesian trajectory and video recordings. PCA components contributing to the 95\% are highlighted in {\setlength{\fboxsep}{1pt}\colorbox{LimeGreen!35}{green}}.}
    \label{tab:result_DoF_full}
    \centering
    \vskip 0.06in
    \setlength{\tabcolsep}{3.8pt}
    \renewcommand{\arraystretch}{0.7}
    \begin{tabular}{lccccccccccc}
    \toprule
    & Mass & Sin. & Dou. & Two & Three & Damp. & Force. & Sin. & Dou. & Sin. & Dou.\\
    & Spring & Pend. & Pend. & Body & Body & Pend. & Duff. & Pend. & Pend. & Pend. & Pend.\\
    \midrule
    Mode & Traj. & Traj. & Traj. & Traj. & Traj. & Traj. & Traj. & Traj. & Traj. & Video & Video\\
    \midrule
    Source & Sim & Sim & Sim & Sim & Sim & Sim & Sim & Real & Real & Real & Real\\
    \midrule
    $\mathrm{PVE}_{1}$ & {\setlength{\fboxsep}{1pt}\colorbox{LimeGreen!35}{98.85}} & {\setlength{\fboxsep}{1pt}\colorbox{LimeGreen!35}{97.16}} & {\setlength{\fboxsep}{1pt}\colorbox{LimeGreen!35}{78.70}} & {\setlength{\fboxsep}{1pt}\colorbox{LimeGreen!35}{95.87}} & {\setlength{\fboxsep}{1pt}\colorbox{LimeGreen!35}{81.87}} & {\setlength{\fboxsep}{1pt}\colorbox{LimeGreen!35}{95.79}} & {\setlength{\fboxsep}{1pt}\colorbox{LimeGreen!35}{96.47}} & {\setlength{\fboxsep}{1pt}\colorbox{LimeGreen!35}{96.28}} & {\setlength{\fboxsep}{1pt}\colorbox{LimeGreen!35}{61.39}} & {\setlength{\fboxsep}{1pt}\colorbox{LimeGreen!35}{98.69}} & {\setlength{\fboxsep}{1pt}\colorbox{LimeGreen!35}{70.47}}\\
    $\mathrm{PVE}_{2}$ & 0.48 & 0.79 & {\setlength{\fboxsep}{1pt}\colorbox{LimeGreen!35}{18.39}} & 1.32 & {\setlength{\fboxsep}{1pt}\colorbox{LimeGreen!35}{17.40}} & 1.28 & 1.02 & 1.91 & {\setlength{\fboxsep}{1pt}\colorbox{LimeGreen!35}{35.41}} & 0.64 & {\setlength{\fboxsep}{1pt}\colorbox{LimeGreen!35}{24.63}}\\
    $\mathrm{PVE}_{3}$ & 0.18 & 0.58 & 0.90 & 1.05 & 0.24 & 0.79 & 0.76 & 0.70 & 1.96 & 0.46 & 0.88\\
    $\mathrm{PVE}_{4}$ & 0.14 & 0.39 & 0.84 & 0.73 & 0.14 & 0.72 & 0.44 & 0.39 & 0.67 & 0.16 & 0.86\\
    $\mathrm{PVE}_{5}$ & 0.11 & 0.34 & 0.59 & 0.52 & 0.12 & 0.50 & 0.35 & 0.22 & 0.30 & 0.12 & 0.67\\
    $\mathrm{PVE}_{6\text{-}10}$ & 0.24 & 0.74 & 0.58 & 0.51 & 0.23 & 0.92 & 0.96 & 0.50 & 0.27 & 0.09 & 1.71\\
    \midrule
    DoF & 1 & 1 & 2 & 1 & 2 & 1 & 1 & 1 & 2 & 1 & 2\\
    \bottomrule
    \end{tabular}
\end{table}
\begin{equation}
    \ddot z = \left(\ddot r-r\dot\phi^{2}\right)e^{\mathrm{i}\phi} + \left(r\ddot\phi+2\dot r\dot\phi\right)\mathrm{i}e^{\mathrm{i}\phi}.
\end{equation}
Equating real and imaginary parts yields
\begin{equation}
    \ddot r-r\dot\phi^{2} = -\frac{\lambda}{r^{2}}, \quad\quad r\ddot\phi+2\dot r\dot\phi = 0.
\end{equation}
Solving for the second derivatives gives the reduced equations of motion
\begin{equation}
    \ddot r = r\dot\phi^{2}-\frac{\lambda}{r^{2}}, \quad\quad \ddot\phi = -2\frac{\dot r\dot\phi}{r}.
\end{equation}
If the reference configuration or units are normalized so that $\lambda=1$, then
\begin{equation}
    \ddot r = r\dot\phi^{2}-\frac{1}{r^{2}}, \quad\quad \ddot\phi = -2\frac{\dot r\dot\phi}{r}.
\end{equation}
Therefore, under the equilateral homographic constraint, the planar three-body system reduces to a two-degree-of-freedom system with generalized coordinates $r$ and $\phi$. The coordinate $r$ describes the changing size of the equilateral triangle, while $\phi$ describes its global rotation.
%
%
\subsection{Identifying Degree of Freedom}\label{apn:results_dof}
We measure the resulting dimensionality by performing PCA~\citep{abdi2010principal} on the latent trajectories and computing the percentage variance explained by each eigenvalue $\lambda_{k}$,
\begin{equation}
    \mathrm{PVE}_{k} = \nicefrac{\lambda_{k}}{\sum \lambda_{k}}.
\end{equation}
The system's degrees of freedom are then given by the smallest number of components whose cumulative PVE exceeds 95\%~\citep{james2021introduction}.
We sample 4,096 random motion sample $\epsilon_{m}$ and encode them to phase space trajectories.
We compute the PCA on these 4,096 phase space trajectories, yielding a 10-component PVE.
We include the full 10-component PCA variance explained metric in Table~\ref{tab:result_DoF_full}.

We further try to understand if the structure of the latent phase space persists through time and initial conditions.
Instead of performing PCA over the entire trajectory, we performance the PCA analysis on points corresponding to $t=\left\{0, 15, 30\right\}$.
We report the absolute cosine similarity between the first two eigenvectors of the phase space of the three body system at $t=\left\{0, 15, 30\right\}$, where Table~\ref{tab:ablation_manifold} shows that the phase space at different times are geometrically equivalent.
Therefore the learned phase space is able to model the full dynamics with a single parameterization.
\begin{table}[t]
    \caption{\textbf{Alignment of phase space at different times.} We report the absolute cosine similarity between the first two eigenvectors of the phase space for the three-body system at $t=\left\{0, 15, 30\right\}$, where the phase space at different times are geometrically equivalent}
    \label{tab:ablation_manifold}
    \centering
    \vskip 0.03in
    \begin{minipage}{0.48\linewidth}
    \raggedleft
    \begin{tabular}{l|ccc}
    \toprule
    $\mathbf{a}_{1}$ & $t=0$ & $t=15$ & $t=30$\\
    \midrule
    $t=0$ & 1.000 & 0.9997 & 0.9988\\
    $t=15$ & 0.9997 & 1.000 & 0.9997\\
    $t=30$ & 0.9988 & 0.9997 & 1.000\\
    \bottomrule
    \end{tabular}
    \end{minipage}
    \hfill
    \begin{minipage}{0.48\linewidth}
    \raggedright
    \begin{tabular}{l|ccc}
    \toprule
    $\mathbf{a}_{2}$ & $t=0$ & $t=15$ & $t=30$\\
    \midrule
    $t=0$ & 1.000 & 0.9992 & 0.9968\\
    $t=15$ & 0.9992 & 1.000 & 0.9993\\
    $t=30$ & 0.9968 & 0.9993 & 1.000\\
    \bottomrule
    \end{tabular}
    \end{minipage}
    \vspace{-0.15in}
\end{table}
%
%
\subsection{Recovering Canonical Coordinates}\label{apn:results_cancor}
We include the recovered canonical coordinates for mass spring and single pendulum in Figure~\ref{fig:SPSGAN_CanonicalCoordinates_Appendix} below.
For the mass spring (left), sweeping the single active direction produces decoded positions that traces a sine wave in time, recovering the verticle displacement $y$ that defines the pendulum's motion.
For the single pendulum (right), sweeping the single active direction produces decoded positions that trace an arc, recovering the angular coordinate $\theta$ that defines the pendulum's motion. 
\begin{figure}[b]
    \vspace{-0.15in}
    \centering
    \begin{subfigure}[t]{0.95\textwidth}
        \centering
        \includegraphics[width=0.52\columnwidth]{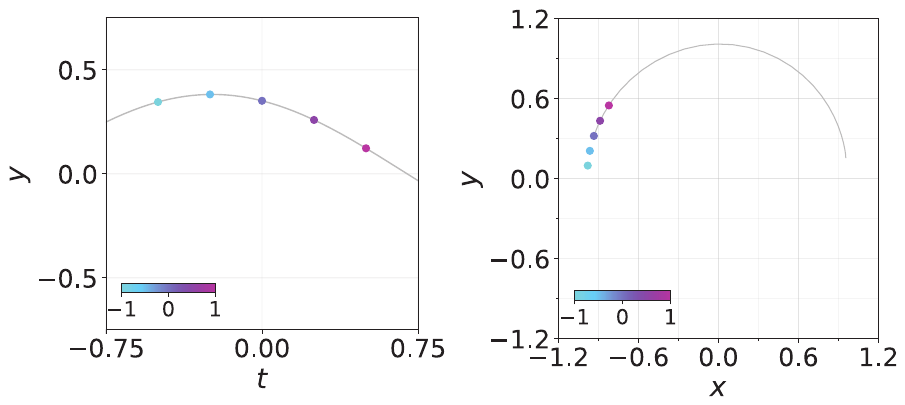}
        \vskip -0.06in
        \caption{\textbf{Recovering system canonical coordinates.} SPS-GAN recovers the correct canonical coordinates, specifically \textbf{(left)} displacement $y$ for mass spring and \textbf{(right)} angular position $\theta$ for single pendulum.}
        \label{fig:SPSGAN_CanonicalCoordinates_Appendix}
    \end{subfigure}
    ~ 
    \begin{subfigure}[t]{0.95\textwidth}
        \centering
        \includegraphics[width=\columnwidth]{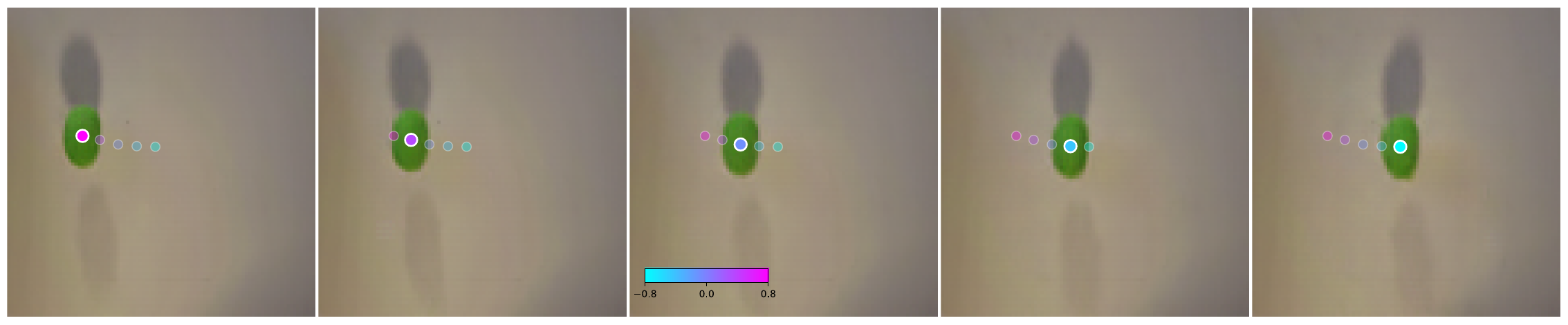}
        \vskip -0.06in
        \caption{\textbf{Recovering system canonical coordinates from video sequences.} SPS-GAN recovers the correct canonical coordinates $\theta$ from videos of a single pendulum~\citep{garcia2025learning}.}
        \label{fig:SPSGAN_CanonicalCoordinatesVideo}
    \end{subfigure}
    \caption{\textbf{Recovering system canonical coordinates.} We show the recovered canonical coordinates for \textbf{(a)} simulated mass-spring and single pendulum, and \textbf{(b)} real world single pendulum videos.}
\end{figure}

Furthermore, perform the same walk on the phase space and decode to video frames in Figure~\ref{fig:SPSGAN_CanonicalCoordinatesVideo}, sweeping the single active direction produces decoded positions that trace an arc, recovering the angular coordinate $\theta$ that defines the pendulum's motion.
This demonstrates that the interpretability and phase-space structure extends between observation modalities.
%
\section{Ablation and Robustness Analysis}\label{apn:ablation_robustness}
\subsection{Robustness of Phase Space Dimension}\label{subsec:ablation_phasespace_dimension}
\begin{table}[t]
    \caption{\textbf{Robustness of phase space dimension.} We train SPS-GAN with phase space with dimension $d_{\mathrm{lat}}=\left\{10,100,1000\right\}$, where SPS-GAN is able to consistently recover a two degrees of freedom representation for the three-body system. Furthermore, increase phase space dimension yields marginal increase in computational cost, both in terms of memory and training time, suggesting that moderate overparameterization is practical. CPVE$_{2}$ denotes the cumulative PVE of the first two components.}
    \label{tab:ablation_dlat}
    \centering
    \vskip 0.03in
    \renewcommand{\arraystretch}{0.85}
    \begin{tabular}{lccc}
    \toprule
    $d_{\mathrm{lat}}$ & 10 & 100 & 1000\\
    $\lambda_{\mathrm{cyclic}}$ & 0.03 & 0.03 & 0.03\\
    \midrule
    DoF & 2 & 2 & 2\\
    CPVE$_{2}$ & $99.27\%$ & $99.16\%$ & $99.20\%$\\
    MSE & $1.26\times10^{-3}$ & $1.49\times10^{-3}$ & $1.75\times10^{-3}$\\
    ep/s & 0.236 & 0.233 & 0.221\\
    MB & 389 & 391 & 433\\
    \bottomrule
    \end{tabular}
    \vspace{-0.15in}
\end{table}
\textbf{Latent-Space Upper Bound.}
SPS-GAN does not assume prior knowledge on the size of the learned phase space.
Therefore, in practice we choose $d_{\mathrm{lat}}$ as an arbitrarily large upper bound on the expected degrees of freedom of the systems being modeled.
We train SPS-GAN with $d_{\mathrm{lat}}=\left\{10,100,1000\right\}$ in Table~\ref{tab:ablation_dlat} on the three-body system.
SPS-GAN is able to consistently recover the correct two degree of freedom representation with similar first two cumulative PVE (CPVE) and MSE.
The exact dimension of the phase space also has minimal effects on the memory requirement and training time, suggesting that moderate overparameterization is computational practical.

\textbf{Sensitivity to the PVE Threshold.}
\begin{table}[b]
    \vspace{-0.15in}
    \caption{\textbf{Sensitivity of active dimension recovery to PVE threshold.} Each entry reports the number of training seeds recovering the expected DoF out of three. The first five rows use simulated trajectories, while the final two use real videos.}
    \label{tab:ablation_pve_threshold}
    \centering
    \vskip 0.03in
    \renewcommand{\arraystretch}{0.85}
    \begin{tabular}{lcccccc}
        \toprule
        System & 90\% & 95\% & 96\% & 97\% & 98\% & 99\%\\
        \midrule
        Mass Spring & 3/3 & 3/3 & 3/3 & 3/3 & 3/3 & 0/3\\
        Single Pend. & 3/3 & 3/3 & 3/3 & 3/3 & 1/3 & 0/3\\
        Double Pend. & 3/3 & 3/3 & 3/3 & 3/3 & 0/3 & 0/3\\
        Two Body & 3/3 & 3/3 & 1/3 & 0/3 & 0/3 & 0/3\\
        Three Body & 3/3 & 3/3 & 3/3 & 3/3 & 3/3 & 3/3\\
        Single Pend. Video & 3/3 & 3/3 & 3/3 & 3/3 & 3/3 & 0/3\\
        Double Pend. Video & 3/3 & 3/3 & 1/3 & 0/3 & 0/3 & 0/3\\
        \bottomrule
    \end{tabular}
\end{table}
We examine whether the recovered dimension depends on the choice of the 95\% cumulative PVE threshold in Table~\ref{tab:ablation_pve_threshold}. 
We vary the threshold from 90\% to 99\% across three training seeds (72, 172, 272) and report how many seeds recover the expected dimension in each setting. 
As shown in Table 10, both 90\% and 95\% recover the expected dimension for all three seeds across five simulated trajectory settings and two real video settings. 
Stricter thresholds increasingly count small residual-variance components as active, particularly for the two-body system and double-pendulum video.
Recovery is therefore consistent over the tested 90–95\% range, but is not invariant to the threshold.

\textbf{Eigenvalue-Ratio Diagnostic.}
We further check whether the recovered dimension corresponds to a clear separation in the PCA spectrum. 
Using the same eigenvalues, we identify the index
\begin{equation}
    \hat{d}_{\mathrm{ratio}}=\arg\max_{k}\frac{\lambda_{k}}{\lambda_{k+1}},
\end{equation}
which marks the largest relative decrease in variance between consecutive components. 
As shown in Table~\ref{tab:ablation_eigenvalue_ratio}, this diagnostic agrees with the expected active dimension in all eleven settings, including non-conservative systems and real trajectory and video observations. 
The agreement corroborates the dimensions obtained with the 95\% criterion without requiring a cumulative variance cutoff.
\begin{table}[t]
    \caption{\textbf{Eigenvalue-ratio diagnostic of the recovered active dimension.} We report ratios $\lambda_k/\lambda_{k+1}$ between consecutive PCA eigenvalues. Highlights the largest ratio, whose index agrees with the expected active dimension derived from PVE threshold as well as the analytical DoF.}
    \label{tab:ablation_eigenvalue_ratio}
    \centering
    \vskip 0.03in
    \renewcommand{\arraystretch}{0.85}
    \begin{tabular}{lllcccc}
        \toprule
        System & Mode & Source & DoF & $k=1$ & $k=2$ & $k=3$\\
        \midrule
        Mass Spring & Traj. & Sim & 1 & {\setlength{\fboxsep}{1pt}\colorbox{LimeGreen!35}{205.9}} & 2.7 & 1.3\\
        Single Pend. & Traj. & Sim & 1 & {\setlength{\fboxsep}{1pt}\colorbox{LimeGreen!35}{123.0}} & 1.4 & 1.5\\
        Double Pend. & Traj. & Sim & 2 & 4.3 & {\setlength{\fboxsep}{1pt}\colorbox{LimeGreen!35}{20.4}} & 1.1\\
        Two Body & Traj. & Sim & 1 & {\setlength{\fboxsep}{1pt}\colorbox{LimeGreen!35}{72.6}} & 1.3 & 1.4\\
        Three Body & Traj. & Sim & 2 & 4.7 & {\setlength{\fboxsep}{1pt}\colorbox{LimeGreen!35}{72.5}} & 1.7\\
        Damped Pend. & Traj. & Sim & 1 & {\setlength{\fboxsep}{1pt}\colorbox{LimeGreen!35}{74.8}} & 1.6 & 1.1\\
        Forced Duffing & Traj. & Sim & 1 & {\setlength{\fboxsep}{1pt}\colorbox{LimeGreen!35}{94.6}} & 1.3 & 1.7\\
        Single Pend. & Traj. & Real & 1 & {\setlength{\fboxsep}{1pt}\colorbox{LimeGreen!35}{50.4}} & 2.7 & 1.8\\
        Double Pend. & Traj. & Real & 2 & 1.7 & {\setlength{\fboxsep}{1pt}\colorbox{LimeGreen!35}{18.1}} & 2.9\\
        Single Pend. & Video & Real & 1 & {\setlength{\fboxsep}{1pt}\colorbox{LimeGreen!35}{154.2}} & 1.4 & 2.9\\
        Double Pend. & Video & Real & 2 & 2.9 & {\setlength{\fboxsep}{1pt}\colorbox{LimeGreen!35}{28.0}} & 1.0\\
        \bottomrule
    \end{tabular}
    \vspace{-0.15in}
\end{table}

\textbf{Inference-Time Pruning of Inactive Components.}
We test whether the identified active subspace retains the information needed for accurate trajectory generation.
At inference, we zero the inactive PCA components and evaluate trajectory MSE across the five conservative systems.
As shown in Table~\ref{tab:ablation_active_dimension_pruning}, the pruned model maintains accuracy close to full SPS-GAN and outperforms the best supervised baseline on four of the five systems.
These results provide functional support for the recovered dimension, as removing the components classified as inactive has only a limited effect on generated motion.
\begin{table}[b]
    \vspace{-0.15in}
    \caption{\textbf{Trajectory accuracy after pruning inactive PCA components.} We report MSE for full SPS-GAN, its inference-time pruned variant, and the best supervised baseline for each system.}
    \label{tab:ablation_active_dimension_pruning}
    \centering
    \vskip 0.03in
    \renewcommand{\arraystretch}{0.7}
    \setlength{\tabcolsep}{4pt}
    \begin{tabular}{lccccc}
        \toprule
        \multirow{2}{*}{Network} & Mass & Single & Double & Two & Three\\
        & Spring & Pend. & Pend. & Body & Body\\
        \midrule
        SPS-GAN & $6.16\times10^{-4}$ & $4.29\times10^{-4}$ & $3.25\times10^{-2}$ & $3.07\times10^{-3}$ & $1.26\times10^{-3}$\\
        SPS-GAN Pruned & $5.79\times10^{-4}$ & $4.25\times10^{-4}$ & $3.54\times10^{-2}$ & $3.27\times10^{-3}$ & $1.33\times10^{-3}$\\
        Best Baseline & $1.13\times10^{-4}$ & $6.27\times10^{-4}$ & $4.66\times10^{-2}$ & $3.61\times10^{-3}$ & $1.37\times10^{-3}$\\
        \bottomrule
    \end{tabular}
\end{table}
\subsection{Ablation Study on Phase Space Dynamics}
Inductive biases have long shown to help neural networks with generalizability and robustness to out-of-distribution shifts~\citep{yang2024learning,zhong2025gagrasp,yang2024behavior,cohen2016group,greydanus2019hamiltonian,cranmer2020lagrangian} on a wide range of machine learning applications. To better understand the role that the port Hamiltonian formulation plays in SPS-GAN, We replacing the phase space dynamics with either GRU~\citep{cho2014learning} or SympNet~\citep{jin2020sympnets}.
The two ablation architectures represents two levels of ablation, where replacing pHNN with SympNet still preserves the symplectic nature of the model, while replacing pHNN with GRU reduces the model down to a black box.

While replacing pHNN with GRU and SympNet are straight forward from a network perspective, careful consideration must be taken to ensure that the change in performance originates solely from the change in dynamics modeling.
The key challenge is how the phase space dynamics interfaces with the cyclic coordinate loss.
Recall that under the conservative generalization of $\hat{D}=\hat{F}=0$, the cyclic coordinate loss for conservative systems~\citep{allen2024hamiltonian} is defined as
\begin{equation}
    \mathcal{L}_{\mathrm{cyclic}} = \lambda_{\mathrm{cyclic}}\sum\nolimits_{i=1}^{d_{\mathrm{lat}}}\left|\dot{p}_{i}\right|.
\end{equation}
The cyclic coordinate loss can be naively implemented using pHNN as it directly outputs derivatives of momentum.
Contrast this to SympNet or GRU that provides state predictions $\left(\mathbf{q}_{t+1}, \mathbf{p}_{t+1}\right)$ directly from current state $\left(\mathbf{q}_{t}, \mathbf{p}_{t}\right)$.
Therefore, implementing the cyclic coordinate loss with both SympNet and GRU requires estimating the time derivatives.
Therefore, we apply the cyclic coordinate loss as
\begin{equation}
    \tilde{\mathcal{L}}_{\mathrm{cyclic}} = \lambda_{\mathrm{cyclic}}\sum\nolimits_{i=1}^{d_{\mathrm{lat}}}\sum\nolimits_{t=0}^{T-1}\left|\frac{p_{i,t+1}-p_{i,t}}{\Delta t}\right|.
\end{equation}
As shown in Table~\ref{tab:ablation}, SPS-GAN with GRU or SympNet phase space dynamcis suffers from both degraded prediction accuracy as well as collapse of interpretability.
\subsection{Ablation Study on Cyclic Coordinate Loss}
\textbf{Sensitivity to Cyclic-Coordinate Loss.}
The dimension of the phase space for SPS-GAN is primiarily regularized through the cyclic coordinate loss, defined by
\begin{equation}
    \mathcal{L}_{\mathrm{cyclic}} = \lambda_{\mathrm{cyclic}}\sum\nolimits_{i=1}^{d_{\mathrm{lat}}}\left|\dot{p}_{i} + \left[\mathbf{D}_{\phi}(\xi)\nabla_{\mathbf{p}
    }\mathcal{H}_{\theta}(\mathbf{z},\xi) - \mathbf{F}_{\psi}\left(\mathbf{u}, \xi\right)\right]_{i}\right|.
\end{equation}
The role of the cyclic coordinate loss is there for to identify the minimally dimension phase space and non-conservative components that is able to accurately model the system dynamics.
The choice of $\lambda_{\mathrm{cyclic}}$ is robust across systems and modalities, as all experiment reported used $\lambda_{\mathrm{cyclic}}=0.03$.
We ablate the cyclic coordinate loss on all five conservative systems by progressively reducing its weight from 0.10 until the loss is removed in Table~\ref{tab:ablation_cyclic}. All tested weights at or above 0.03 recover the expected active dimension with similar MSE, while lower weights can overestimate the active dimension and degrade trajectory accuracy.
\begin{table}[t]
    \caption{\textbf{Sensitivity to cyclic-coordinate loss weight across conservative systems.} We report recovered degrees of freedom and trajectory MSE. All tested weights at or above $\lambda_{\mathrm{cyclic}}=0.03$ recover the expected active dimension for all five systems.}
    \label{tab:ablation_cyclic}
    \centering
    \vskip 0.03in
    \renewcommand{\arraystretch}{0.8}
    \begin{tabular}{lcccccc}
        \toprule
        & \multirow{2}{*}{$\lambda_{\mathrm{cyclic}}$} & Mass & Single & Double & Two & Three\\
        & & Spring & Pend. & Pend. & Body & Body\\
        \midrule
        & $0.10$ & 1 & 1 & 2 & 1 & 2\\
        & $0.05$ & 1 & 1 & 2 & 1 & 2\\
        DoF & $0.03$ & 1 & 1 & 2 & 1 & 2\\
        & $0.01$ & 2 & 2 & 3 & 2 & 2\\
        & $0.00$ & 4 & 5 & 9 & 3 & 4\\
        \midrule
        & $0.10$ & $6.0\times10^{-4}$ & $4.6\times10^{-4}$ & $3.7\times10^{-2}$ & $3.3\times10^{-3}$ & $1.4\times10^{-3}$\\
        & $0.05$ & $6.4\times10^{-4}$ & $4.8\times10^{-4}$ & $3.2\times10^{-2}$ & $2.9\times10^{-3}$ & $1.1\times10^{-3}$\\
        MSE & $0.03$ & $6.2\times10^{-4}$ & $4.3\times10^{-4}$ & $3.3\times10^{-2}$ & $3.1\times10^{-3}$ & $1.3\times10^{-3}$\\
        & $0.01$ & $8.6\times10^{-4}$ & $4.7\times10^{-4}$ & $4.0\times10^{-2}$ & $3.8\times10^{-3}$ & $1.7\times10^{-3}$\\
        & $0.00$ & $8.9\times10^{-4}$ & $6.9\times10^{-4}$ & $6.9\times10^{-2}$ & $4.4\times10^{-2}$ & $4.4\times10^{-3}$\\
        \bottomrule
    \end{tabular}
    \vspace{-0.15in}
\end{table}
\begin{table}[b]
    \vspace{-0.15in}
    \caption{\textbf{Sensitivity to non-conservative regularization weights.} We jointly vary $\lambda_{D}=\lambda_{F}$ on the damped pendulum and report recovered degrees of freedom and trajectory MSE.}
    \label{tab:ablation_nonconservative_weights}
    \centering
    \vskip 0.03in
    \renewcommand{\arraystretch}{0.8}
    \begin{tabular}{lcc}
    \toprule
    $\lambda_{D}=\lambda_{F}$ & DoF & MSE\\
    \midrule
    $0.1$ & 1 & $4.7\times10^{-4}$\\
    $0.01$ & 1 & $4.6\times10^{-4}$\\
    $0.001$ & 1 & $4.9\times10^{-4}$\\
    $0$ & 3 & $2.0\times10^{-3}$\\
    \bottomrule
    \end{tabular}
\end{table}

\textbf{Sensitivity to Non-Conservative Regularization.}
We jointly vary the dissipation and forcing regularization weights on the damped pendulum, setting $\lambda_{D}=\lambda_{F}$ (all other experiments in the paper uses $\lambda_{F}=0.01$). 
As shown in Table~\ref{tab:ablation_nonconservative_weights}, every tested nonzero weight recovers the expected one degree of freedom with similar trajectory MSE across two orders of magnitude.
Removing both penalties increases the recovered dimension to three and substantially increases MSE.
The shared value of 0.01 used for both weights throughout the dissipative and forced experiments lies within this stable range.
\subsection{Ablation Study on Phase Space Map}
\begin{table}[t]
    \caption{\textbf{Robustness to observation noise.} We report the cumulative PVE of the first two components (CPVE$_{2}$) for double-pendulum videos across three training seeds. An SNR of $\infty$ denotes no added noise.}
    \label{tab:ablation_observation_noise}
    \centering
    \vskip 0.03in
    \renewcommand{\arraystretch}{0.85}
    \begin{tabular}{lccc}
        \toprule
        SNR (dB) & Seed 72 & Seed 172 & Seed 272\\
        \midrule
        $\infty$ & $95.10\%$ & $95.21\%$ & $96.33\%$\\
        $30$ & $95.47\%$ & $95.77\%$ & $95.97\%$\\
        $20$ & $95.20\%$ & $96.27\%$ & $95.60\%$\\
        $10$ & $95.33\%$ & $95.13\%$ & $95.45\%$\\
        $0$ & $86.10\%$ & $82.07\%$ & $81.41\%$\\
        \bottomrule
    \end{tabular}
    \vspace{-0.15in}
\end{table}
The phase space map projects the random motion samples $\varepsilon_{\mathrm{m}}$ onto the phase space to generate the initial conditions for the rollout.
This allows the port Hamiltonian formulation to originate from coordinates that are well condition (and in the ideal setting on the motion manifold of the system).
We verify this claim by ablating the phase space map, where SPS-GAN is no longer able to identify a minimal phase space matching system degrees of freedom (see Table~\ref{tab:ablation}).
\subsection{Robustness to Observation Noise}\label{subsec:ablation_observation_noise}
We evaluate the robustness of active dimension recovery by adding per-pixel i.i.d. Gaussian noise to normalized double-pendulum videos.
We retrain SPS-GAN for 20,000 steps at each signal-to-noise ratio (SNR) using three training seeds (72, 172, 272).
As shown in Table~\ref{tab:ablation_observation_noise}, all three seeds recover the expected two degrees of freedom at the tested SNRs down to 10~dB.
At 0~dB, where signal and noise have equal power, the cumulative PVE of the first two components falls below 95\% for every seed, causing the recovered dimension to exceed the expected two degrees of freedom.
These results demonstrate robustness over the tested 10--30~dB range and identify a breakdown regime at 0~dB.
\subsection{Training Stability and Initial-Condition Coverage}\label{subsec:ablation_stability_coverage}
\begin{table}[b]
    \vspace{-0.15in}
    \caption{\textbf{Training stability on real double-pendulum videos.} We report cumulative PVE of the first two components and FVD at selected training checkpoints.}
    \label{tab:ablation_training_stability}
    \centering
    \vskip 0.03in
    \renewcommand{\arraystretch}{0.8}
    \begin{tabular}{lcccccc}
    \toprule
    Training step & 10k & 12k & 14k & 16k & 18k & 20k\\
    \midrule
    CPVE$_{2}$ (\%) & 95.39 & 95.17 & 95.75 & 96.15 & 95.19 & 95.10\\
    FVD & 194.75 & 190.90 & 152.66 & 88.41 & 78.06 & 74.69\\
    \bottomrule
    \end{tabular}
\end{table}
\textbf{Training stability.} 
We evaluate whether the recovered active dimension remains stable as training progresses on real double-pendulum videos in Table~\ref{tab:ablation_training_stability}. 
The 95\% cumulative PVE criterion recovers the expected two degrees of freedom at 102 of 106 checkpoints from step 9,300 onward, with the four exceptions confined to steps 9,400–10,400. 
At the checkpoints reported below, the cumulative PVE of the first two components remains above 95\% while FVD decreases, indicating that video quality continues to improve while the recovered dimension remains unchanged.
The active PCA directions also remain aligned across steps 10,000, 15,000, and 20,000, with cosine similarities of 0.97–0.99.

\textbf{Initial-condition coverage.} 
We assess whether generated initial conditions cover the underlying distribution across the five conservative systems in Table~\ref{tab:ablation_initial_condition_coverage}. 
We compare generated and reference initial conditions using forward KL divergence and sliced Wasserstein distance (SWD), with real-to-real comparisons providing a reference floor for each metric. 
Both distances remain within a small factor of their respective floors across all five systems. 
These results support broad coverage of the initial-condition distributions and provide evidence against mode collapse in the evaluated settings.
\begin{table}[t]
    \caption{\textbf{Distributional coverage of generated initial conditions.} We report forward KL divergence and SWD between generated and reference initial conditions, alongside their real-to-real reference floors.}
    \label{tab:ablation_initial_condition_coverage}
    \centering
    \vskip 0.03in
    \renewcommand{\arraystretch}{0.8}
    \begin{tabular}{lccccc}
    \toprule
    \multirow{2}{*}{Metric} & Mass & Single & Double & Two & Three\\
    & Spring & Pend. & Pend. & Body & Body\\
    \midrule
    KL floor & 0.05 & 0.04 & 0.44 & 0.27 & 0.32\\
    KL & 0.07 & 0.06 & 0.99 & 0.41 & 0.75\\
    SWD floor & 0.073 & 0.080 & 0.085 & 0.092 & 0.109\\
    SWD & 0.090 & 0.093 & 0.121 & 0.107 & 0.117\\
    \bottomrule
    \end{tabular}
    \vspace{-0.15in}
\end{table}
%
\section{Experiment Statistical Variance}\label{apn:exp_var}
In this section we report the statistical variance for SPS-GAN and baselines methods for experiments reported in Table~\ref{tab:result_conservative_accuracy} and Table~\ref{tab:result_nonconservative_accuracy} in Table~\ref{tab:apn_conservative_var} and Table~\ref{tab:apn_nonconservative_var} respectively.
We calculate the standard deviation of the results reported based on three training seeds (72, 172, 272), using the same original training data.
We calculate the mean and standard deviation of the MSE such that
\begin{equation}
    \bar{\mu} = \frac{1}{n}\sum\nolimits_{i=1}^{3} \mu_{i}, \quad\quad \sigma = \sqrt{\frac{1}{n-1}\sum\nolimits_{i=1}^{3}(\mu_{i}-\bar{\mu})^{2}}.
\end{equation}
%
\section{Broader Impact}
SPS-GAN may support scientific discovery and engineering by learning compact, interpretable representations of observed dynamics without prescribed state variables or complete analytical models. Its generated trajectories and videos could aid hypothesis exploration and synthetic data generation where physical experiments are costly or difficult to repeat.

Physically accurate generative modeling could also be misused to make fabricated depictions of physical events more convincing. Physically plausible outputs may further be mistaken for validated predictions, since mechanical structure does not guarantee accuracy outside the observed conditions. While we do not believe SPS-GAN meaningfully changes the risk landscape relative to current generative methods, generated data should be distinguished from measurements and validated against independent observations before informing consequential scientific or engineering decisions.
\begin{table}[h]
    \caption{\textbf{Trajectory generation accuracy on conservative systems.} The MSE of SPS-GAN generated trajectory is compared to predicted trajectory from baselines. SPS-GAN models single systems, while cSPS-GAN jointly models all five conditionally. When generating single system dynamics, unsupervised SPS-GAN outperforms supervised baselines. When jointly modeling systems, supervised conditional cSPS-GAN performs on par with single system baselines.}
    \label{tab:apn_conservative_var}
    \renewcommand{\arraystretch}{0.8}
    \centering
    \vskip 0.03in
    \begin{tabular}{lccc}
    \toprule
    Network & Mass Spring & Single Pend. & Double Pend.\\
    \midrule
    SPS-GAN & $\left(6.16\pm0.34\right)\times10^{-4}$ & $\boldsymbol{\left(4.29\pm0.64\right)\times10^{-4}}$ & $\boldsymbol{\left(3.25\pm0.14\right)\times10^{-2}}$\\
    HNN & $\left(3.53\pm0.15\right)\times10^{-4}$ & $\left(6.27\pm0.32\right)\times10^{-4}$ & $\left(1.25\pm0.89\right)\times10^{-1}$\\
    SymODEN & $\left(6.40\pm0.36\right)\times10^{-4}$ & $\left(6.56\pm0.45\right)\times10^{-4}$ & $\left(9.89\pm0.51\right)\times10^{-2}$\\
    TRS-ODEN & $\left(6.72\pm0.68\right)\times10^{-4}$ & $\left(9.57\pm1.87\right)\times10^{-4}$ & $\left(6.12\pm0.31\right)\times10^{-2}$\\
    NeuralODE & $\left(1.62\pm0.60\right)\times10^{-4}$ & $\left(1.59\pm1.22\right)\times10^{-3}$ & $\left(5.99\pm1.32\right)\times10^{-2}$\\
    LatentODE & $\boldsymbol{\left(1.13\pm0.03\right)\times10^{-4}}$ & $\left(9.27\pm1.06\right)\times10^{-4}$ & $\left(4.66\pm0.49\right)\times10^{-2}$\\
    \midrule
    cSPS-GAN & $\left(9.88\pm0.32\right)\times10^{-4}$ & $\left(5.11\pm0.25\right)\times10^{-4}$ & $\left(3.33\pm0.63\right)\times10^{-2}$\\
    \bottomrule
    \end{tabular}
    \vskip 0.03in
    \begin{tabular}{lcc}
    \toprule
    Network & Two Body & Three Body\\
    \midrule
    SPS-GAN & $\boldsymbol{\left(3.07\pm0.15\right)\times10^{-3}}$ & $\boldsymbol{\left(1.26\pm0.04\right)\times10^{-3}}$\\
    HNN &  $\left(3.61\pm0.18\right)\times10^{-3}$ & $\left(2.81\pm0.08\right)\times10^{-3}$\\
    SymODEN & $\left(6.20\pm0.25\right)\times10^{-3}$ & $\left(8.08\pm2.85\right)\times10^{-3}$\\
    TRS-ODEN & $\left(6.36\pm1.13\right)\times10^{-3}$ & $\left(6.22\pm0.36\right)\times10^{-3}$\\
    NeuralODE & $\left(5.72\pm0.37\right)\times10^{-3}$ & $\left(3.67\pm0.57\right)\times10^{-3}$\\
    LatentODE & $\left(1.98\pm0.98\right)\times10^{-2}$ & $\left(1.37\pm0.10\right)\times10^{-3}$\\
    \midrule
    cSPS-GAN & $\left(4.46\pm0.17\right)\times10^{-3}$ & $\left(2.61\pm0.11\right)\times10^{-3}$\\
    \bottomrule
    \end{tabular}
\end{table}
\begin{table}[h]
    \caption{\textbf{Trajectory generation accuracy on non-conservative systems.} The MSE error of SPS-GAN generated trajectory is reported compared to predicted trajectory from baselines. Unsupervised SPS-GAN outperforms supervised single-system baselines.}
    \label{tab:apn_nonconservative_var}
    \renewcommand{\arraystretch}{0.8}
    \centering
    \vskip 0.03in
    \begin{tabular}{lccc}
    \toprule
    Network & Ideal Pend. & Damped Pend. & Forced Duffing\\
    \midrule
    SPS-GAN & $\boldsymbol{\left(4.77\pm0.13\right)\times10^{-4}}$ & $\boldsymbol{\left(4.57\pm0.06\right)\times10^{-4}}$ & $\boldsymbol{\left(5.22\pm0.62\right)\times10^{-4}}$\\
    pHNN & $\left(8.42\pm0.88\right)\times10^{-4}$ & $\left(4.76\pm0.19\right)\times10^{-4}$ & $\left(6.92\pm0.84\right)\times10^{-4}$\\
    NeuralODE & $\left(1.59\pm1.22\right)\times10^{-3}$ & $\left(6.85\pm0.57\right)\times10^{-4}$ & $\left(7.35\pm0.59\right)\times10^{-4}$\\
    LatentODE & $\left(9.27\pm1.06\right)\times10^{-4}$ & $\left(6.69\pm0.27\right)\times10^{-4}$ & $\left(1.18\pm0.31\right)\times10^{-3}$\\
    TRS-ODEN & $\left(9.57\pm1.87\right)\times10^{-4}$ & $\left(1.44\pm0.32\right)\times10^{-3}$ & $\left(6.50\pm0.38\right)\times10^{-4}$\\
    \bottomrule
    \end{tabular}
\end{table}
%
\clearpage
\section*{NeurIPS Paper Checklist}

\begin{enumerate}

\item {\bf Claims}
    \item[] Question: Do the main claims made in the abstract and introduction accurately reflect the paper's contributions and scope?
    \item[] Answer: \answerYes{} 
    \item[] Justification: The paper claims four key contributions in the introduction, namely:
    \begin{enumerate}
        \itemsep0em 
        \item Generates physically consistent rollouts across multiple systems, parameter variations, and observation modalities (Cartesian coordinates and video);
        \item Adopts a port-Hamiltonian formulation to model conservative, dissipative, and forced systems within a single framework;
        \item Identifies the degrees of freedom of a system from observations alone, through a cyclic-coordinate regularizer that compresses the phase space to its minimally active dimensions;
        \item Recovers physically meaningful canonical coordinates by decoding along the active phase space dimensions, connecting learned representations to underlying physical quantities.
    \end{enumerate}
    We show that
    \begin{enumerate}
        \itemsep0em 
        \item SPS-GAN achieves improved dynamical accuracy in Table~\ref{tab:result_conservative_accuracy}, \ref{tab:result_nonconservative_accuracy}, and \ref{tab:result_single_conservative_video}, where SPS-GAN outperforms competitive baselines;
        \item SPS-GAN extends to non-conservative systems using simulated damped pendulum and forced duffing, as well as real world single and double pendulum, both in terms of trajectory and video, where SPS-GAN is able to faithfully recover both the degrees of freedom as well as the canonical coordinates;
        \item SPS-GAN correctly identifies the system degrees of freedom over 11 systems in Table~\ref{tab:result_DoF};
        \item SPS-GAN correctly identifies the system canonical coordinates for 6 systems in Figure~\ref{fig:SPSGAN_CanonicalCoordinates}, \ref{fig:SPSGAN_CanonicalCoordinates_Appendix}, and \ref{fig:SPSGAN_CanonicalCoordinatesVideo};
    \end{enumerate}

\item {\bf Limitations}
    \item[] Question: Does the paper discuss the limitations of the work performed by the authors?
    \item[] Answer: \answerYes{}
    \item[] Justification: We have included limitations after the conclusion of the paper, where we highlight that SPS-GAN can only accommodate systems that are Hamiltonian or port-Hamiltonian, as well as videos that exhibit approximately planar moiton.

\item {\bf Theory assumptions and proofs}
    \item[] Question: For each theoretical result, does the paper provide the full set of assumptions and a complete (and correct) proof?
    \item[] Answer: \answerYes{}
    \item[] Justification: We include detailed derivations of the system degrees of freedom and canonical coordinates in Appendix~\ref{subsec:DoF_canonical_coordinates}.

    \item {\bf Experimental result reproducibility}
    \item[] Question: Does the paper fully disclose all the information needed to reproduce the main experimental results of the paper to the extent that it affects the main claims and/or conclusions of the paper (regardless of whether the code and data are provided or not)?
    \item[] Answer: \answerYes{}
    \item[] Justification: We report the experimental setting and hyper-parameters in Appendix~\ref{apn:dataset}. The training and inference code is open-sourced in the GitHub link below.

\item {\bf Open access to data and code}
    \item[] Question: Does the paper provide open access to the data and code, with sufficient instructions to faithfully reproduce the main experimental results, as described in supplemental material?
    \item[] Answer: \answerYes{}
    \item[] Justification: Open source code is provided in the following GitHub repo: \url{https://github.com/yy0555/SPS-GAN}.

\item {\bf Experimental setting/details}
    \item[] Question: Does the paper specify all the training and test details (e.g., data splits, hyperparameters, how they were chosen, type of optimizer) necessary to understand the results?
    \item[] Answer: \answerYes{}
    \item[] Justification: We include all training details in Appendix~\ref{apn:dataset} as well as data generation procedure and governing equations. All experiments that use existing datasets are cited when first introduced.

\item {\bf Experiment statistical significance}
    \item[] Question: Does the paper report error bars suitably and correctly defined or other appropriate information about the statistical significance of the experiments?
    \item[] Answer: \answerYes{}
    \item[] Justification: We include standard deviations for the numerical results in Appendix~\ref{apn:exp_var}. We do not include error statistics for qualitative results such as decoded canonical coordinates.

\item {\bf Experiments compute resources}
    \item[] Question: For each experiment, does the paper provide sufficient information on the computer resources (type of compute workers, memory, time of execution) needed to reproduce the experiments?
    \item[] Answer: \answerYes{}
    \item[] Justification: Yes include the compute requirement in Appendix~\ref{apn:dataset}. Specifically, all training of SPS-GAN was performed on cluster nodes each with 1 Nvidia L40 GPU, 16 core partition of an Intel Xeon Gold 5320, and 64G of DDR4 3200MHz RDIMM.
    
\item {\bf Code of ethics}
    \item[] Question: Does the research conducted in the paper conform, in every respect, with the NeurIPS Code of Ethics \url{https://neurips.cc/public/EthicsGuidelines}?
    \item[] Answer: \answerYes{}
    \item[] Justification: The authors have reviewed the code of ethics and identified no deviations. We acknowledge the general harm that generative video models can do, but we do not believe that SPS-GAN poses increased risk compared to contemporary models.

\item {\bf Broader impacts}
    \item[] Question: Does the paper discuss both potential positive societal impacts and negative societal impacts of the work performed?
    \item[] Answer: \answerYes{}
    \item[] Justification: Due to page limitation, we call-out that physically accuracy generative modeling may contribute to malicious use of genAI in the appendix.
    
\item {\bf Safeguards}
    \item[] Question: Does the paper describe safeguards that have been put in place for responsible release of data or models that have a high risk for misuse (e.g., pre-trained language models, image generators, or scraped datasets)?
    \item[] Answer: \answerYes{}
    \item[] Justification: We do not release data in this paper. We will include additional discussion when the github repo is made public.

\item {\bf Licenses for existing assets}
    \item[] Question: Are the creators or original owners of assets (e.g., code, data, models), used in the paper, properly credited and are the license and terms of use explicitly mentioned and properly respected?
    \item[] Answer: \answerYes{}
    \item[] Justification: All existing datasets are cited when first appearing in the paper. All datasets used are according to their release copy right.

\item {\bf New assets}
    \item[] Question: Are new assets introduced in the paper well documented and is the documentation provided alongside the assets?
    \item[] Answer: \answerNA{}
    \item[] Justification: this paper does not release new assets.

\item {\bf Crowdsourcing and research with human subjects}
    \item[] Question: For crowdsourcing experiments and research with human subjects, does the paper include the full text of instructions given to participants and screenshots, if applicable, as well as details about compensation (if any)? 
    \item[] Answer: \answerNA{}
    \item[] Justification: This paper does not contain human subjects.

\item {\bf Institutional review board (IRB) approvals or equivalent for research with human subjects}
    \item[] Question: Does the paper describe potential risks incurred by study participants, whether such risks were disclosed to the subjects, and whether Institutional Review Board (IRB) approvals (or an equivalent approval/review based on the requirements of your country or institution) were obtained?
    \item[] Answer: \answerNA{}
    \item[] Justification: This paper is not subject to IRB as we do not study live subjects.

\item {\bf Declaration of LLM usage}
    \item[] Question: Does the paper describe the usage of LLMs if it is an important, original, or non-standard component of the core methods in this research? Note that if the LLM is used only for writing, editing, or formatting purposes and does \emph{not} impact the core methodology, scientific rigor, or originality of the research, declaration is not required.
    \item[] Answer: \answerNA{} 
    \item[] Justification: LLM is not included in the core contribution of the paper.
\end{enumerate}
\end{document}